\documentclass{elsarticle} 
\usepackage{times}

\usepackage{amsmath,amsfonts,bm}

\def\eqref#1{equation~\ref{#1}}

\def\1{\bm{1}}

\DeclareMathAlphabet{\mathsfit}{\encodingdefault}{\sfdefault}{m}{sl}
\SetMathAlphabet{\mathsfit}{bold}{\encodingdefault}{\sfdefault}{bx}{n}

\usepackage{url}
\usepackage{graphicx}
\usepackage{graphics}
\usepackage{subcaption}
\usepackage{wrapfig}
\usepackage{amssymb}
\usepackage{xcolor}
\usepackage[colorlinks,urlcolor=blue,bookmarks=false,hypertexnames=true]{hyperref}

\newcommand{\bR}{\mathbb{R}}
\newcommand{\cA}{\mathcal{A}}

\newcommand{\cS}{\mathcal{S}}
\newcommand{\cT}{\mathcal{T}}
\newcommand{\cO}{\mathcal{O}}

\usepackage{titlesec}
\titlespacing\section{0pt}{10pt plus 4pt minus 2pt}{3pt plus 2pt minus 2pt}
\titlespacing\subsection{0pt}{10pt plus 4pt minus 2pt}{3pt plus 2pt minus 2pt}

\usepackage[normalem]{ulem}

\title{Negotiating Team Formation Using \\ Deep Reinforcement Learning}

\author{Yoram Bachrach, Richard Everett, Edward Hughes, Angeliki Lazaridou,\\ Joel Z. Leibo, Marc Lanctot, Michael Johanson, Wojciech M. Czarnecki, Thore Graepel \\
DeepMind, London \\ 
\{yorambac, reverett, edwardhughes, angeliki, jzl, lanctot, mjohanson, lejlot, thore\} @google.com
}

\begin{document}

\begin{abstract}
When autonomous agents interact in the same environment, they must often cooperate to achieve their goals. One way for agents to cooperate effectively is to form a team, make a binding agreement on a joint plan, and execute it. However, when agents are self-interested, the gains from team formation must be allocated appropriately to incentivize agreement. Various approaches for multi-agent negotiation have been proposed, but typically only work for particular negotiation protocols. More general methods usually require human input or domain-specific data, and so do not scale. To address this, we propose a framework for training agents to negotiate and form teams using deep reinforcement learning. Importantly, our method makes no assumptions about the specific negotiation protocol, and is instead completely experience driven. We evaluate our approach on both non-spatial and spatially extended team-formation negotiation environments, demonstrating that our agents beat hand-crafted bots and reach negotiation outcomes consistent with fair solutions predicted by cooperative game theory. Additionally, we investigate how the physical location of agents influences negotiation outcomes.
\end{abstract}

\maketitle

\section{Introduction and Related Work}
\label{l_sect_intro}
The ability of humans to cooperate with one another at scale is crucial to our success as a species~\cite{bowles2003origins,rand2013human,harari2014sapiens,henrich2017secret}. We are capable of cooperating with others even when we have somewhat different preferences or when our incentives are only partially aligned~\cite{johnson1989cooperation,komorita1995interpersonal}. Many tasks we face are only efficiently accomplished by a team, ranging from scoring points in a basketball game, through manufacturing products on an assembly line to building successful companies. 

Similarly, multiple artificial agents inhabiting the same environment affect each other, and may gain by coordinating their actions. Indeed, many tasks are effectively intractable for any single agent, and require a team of collaborators. Examples include search and rescue~\cite{kitano1999robocup}, multi-robot patrolling~\cite{agmon2008multi}, security~\cite{tambe2011security}, managing supply chains~\cite{christopher2016logistics} and multiplayer first-person video games~\cite{jaderberg2018human}. 

Despite the need to cooperate, stakeholders sometimes have different abilities and preferences which affect the chosen course of action. Further, although stakeholders benefit from cooperation, they might also be in {\it competition} with one another, possibly vying to  obtain a larger share of the gains. For instance, founders of companies all want the company to succeed, but each would also prefer to receive a larger share of the profit. 

Agents must therefore negotiate to form teams that are both fairly aligned with individual interests and capable of achieving the task at hand. In many settings agents have a choice of partners they can work with. For example, there may be multiple suppliers providing goods required to make an end product, or multiple potential partners to share data with. Hence, agents must reason about who to join forces with to solve a task, and how to split the gains. 

This problem can characterized as a {\em team-formation negotiation} problem as follows~\cite{kraus1997negotiation,shehory1998methods}. By definition, no single agent can perform the task on their own, but there may be several teams of agents who are capable of doing so, so each agent must decide who to collaborate with. The reward for accomplishing the task is awarded to the first team that solves it. 
\footnote{We take a cooperative game approach with binding contracts, rather than examining deviations by agent subsets as done in previous work on strong equilibria or strong Price of Anarchy~\cite{maskin1978implementation,holzman1997strong,andelman2009strong,bachrach2014strong}.
}
Hence, agents 
interact with one another to simultaneously form a team and agree on how to share the joint reward. To solve this abstract problem, one must provide a concrete environment where agents can negotiate and reach an agreement; 
We must specify a \textit{negotiation protocol} that encodes the allowed negotiation actions and determines the agreement reached~\cite{rosenschein1994rules}. 

Team-formation negotiation tasks are natural objects of study in game theory. More precisely, {\em cooperative} game theory focuses on interactions between agents who form teams and make enforceable agreements about outcomes~\cite{brandenburger2007cooperative,chalkiadakis2011computational}. 
\footnote{This setting is akin to {\it ad hoc teamwork}~\cite{Stone10AdHocTeamwork}, except it has an initial phase where agents negotiate before solving the problem together, and make binding agreements about sharing a joint reward.}
{\em Weighted voting games} are an archetypal problem, in which every agent has a weight and a team of agents is successful if the sum of the weights of its participants exceeds a fixed threshold~\cite{banzhaf1964weighted,holler1982forming}. Weighted voting games also offer a simple model of coalition formation in legislative bodies~\cite{leech2002designing,felsenthal1998measurement}.

{\bf Cooperative game theory:} seeks to predict the agreements negotiated by agents in such settings, proposing several solution concepts. Some solutions, such as the core, bargaining set and nucleolus~\cite{aumann1961core,aumann1961bargaining,schmeidler1969nucleolus,aumann1992handbook}, have focused on identifying stable agreements. Other solutions, known as {\em power indices}, have tried to measure the objective negotiation position of agents,  quantifying their relative ability to affect the outcome of the game, or the {\em fair} share of the joint reward they should receive~\cite{chalkiadakis2011computational}. The most prominent of these is the Shapley value~\cite{shapley1953value} which has been widely studied for weighted voting games~\cite{shapley1954method,straffin1988shapley}. In particular, it has been used to estimate political power~\cite{leech2002designing,felsenthal1998measurement,aziz2011false,zuckerman2012manipulating}. 
\footnote{Cooperative game theory has also been applied to analyze many other forms of multi-agent interactions (see recent textbooks for a complete survey~\cite{chalkiadakis2011computational}), including analyzing negotiation power~\cite{kohli1989cooperative,aziz2010monotone,meir2010minimal,aminadav2011rebuilding,bachrach2013computing,procaccia2014structure}, finding key contributors in joint tasks~\cite{conitzer2004computing,ieong2005marginal,bachrach2012crowd,bachrach2012solving,herbrich2016pricing,banarse2019body}, analyzing collusion in auctions~\cite{mailath1991collusion,bachrach2011cooperative,bachrach2010honor,van2012efficiency,lev2013mergers}, identifying critical components in communication or transportation networks~\cite{suris2007cooperative,bachrach2008power,aziz2009power,resnick2009cost,bachrach2009power,bachrach2010path}, characterizing electricity and power systems~\cite{abapour2020game} and even analyzing workloads in cloud computing or Bitcoin mining~\cite{blocq2014shared,lewenberg2015bitcoin} and car-pooling~\cite{li2016dynamic}.
}

{\bf The relation between the Shapley value and Nash equilibria:} Cooperative game theory only considers the value each subset of agents can achieve, and abstracts away many details regarding the specific actions and negotiation protocols agents may use. Such finer-grained analysis can theoretically be achieved by applying non-cooperative game theory, and in particular the Nash equilibrium. Researchers have already noted several connections between the Shapley value and the core and competitive equilibria or Nash equilibria in restricted bargaining and market settings or corporate structures~\cite{gul1989bargaining,zwiebel1995block}. When a certain class of competitive markets is modelled as a multiplayer game, the Shapley value converges to a competitive equilibrium when the set of traders is expanded homogeneously~\cite{shapley1964values}. Further, in markets where each trader only holds a negligible fraction of the resources of the economy (modelled as a game with a continuum of traders), the allocations given by the Shapley value are the same as those obtained under competitive equilibrium~\cite{aumann1975values,aumann2015values}. Additionally, results on the existence of a competitive Walrasian equilibria can be derived from Nash's theorem about the existence of Nash equilibria by applying a cooperative game analysis on Shapley and Shubik's trading-post game~\cite{dubey2003nash}. While these results hold for various market domains, they do not cover arbitrary negotiation settings. 
\footnote{We also analyze the relation between the Shapley value and the Nash equilibrium in the negotiation environments we propose, see Section~\ref{l_shapley_vs_nash}.}
A key question is thus whether learning under a specific negotiation protocols, typically modelled as a non-cooperative game and analyzed through Nash equilibria, is likely to result in outcomes similar to those predicted by cooperative game theory.

{\bf Constructing a Negotiation Agent by Reinforcement Learning:}
Pragmatically speaking, using a multi-agent prism, how should one construct a negotiating agent that maximizes the reward obtained? Many researchers have borrowed ideas from cooperative game theory to hand-craft bots~\cite{zlotkin1989negotiation,aknine2004extended,ito2007multi,bachrach2011rip}, often requiring additional human data~\cite{oliver1996machine,lin2010can}. Such bots are tailored to specific negotiation protocols, so modifying the protocol or switching to a different protocol requires manually re-writing the bot~\cite{jennings2001automated,an2010automated}. As a result, algorithms based purely on cooperative game theory are neither generally applicable nor scalable.

Moreover, negotiation and team formation in the real world is significantly more complex than in the game theoretic setting, for several reasons: (1) negotiation protocols can be arbitrarily complicated and are rarely fixed; (2) enacting a negotiation requires a temporally extended policy; (3) the idiosyncrasies of the environment affect the negotiation mechanics; (4) Players must make decisions based on incomplete information about others' policies.

\subsection{Our Contribution} 
\label{l_sect_our_contrib}

We propose using multi-agent reinforcement learning (RL) as a paradigm for negotiating team formation, which can be applied to arbitrary negotiation protocols in complex environments. Here, individual agents must learn how to solve team formation tasks based on their experiences interacting with others, rather than via hand-crafted algorithms. Our RL approach is automatically applicable to Markov games~\cite{Shapley53,Littman94markovgames}, which are temporally and spatially extended, similar to recent work in the non-cooperative case~\cite{leibo2017multi,LererP17,Foerster17LOLA}. As such, our method falls in the very broad category of multi-agent simulation based methods~\cite{nikolai2009tools,uhrmacher2009multi,kravari2015survey} (see various textbooks for a general introduction to multi-agent based simulation~\cite{ferber1999multi,wooldridge2009introduction}). 
\footnote{Some multi-agent simulation approaches attempt to use formal verification to investigate whether the entire system exhibits desired behavior~\cite{fisher1997formal,xiang2005verification}, whereas others focus on the relation between multi-agent simulations and game theoretic solutions~\cite{SLB09,vorobeychik2008stochastic,shoham2003multi,lanctot2017unified}. This work falls in the second category.}
In contrast to earlier work on multi-agent RL in non-cooperative games, the key novelty of our work is comparing the behaviour of negotiating RL agents with solutions from  {\em cooperative game theory}.

Some previous work in multi-agent (deep) reinforcement learning for negotiation has cast the problem as one of communication, rather than team formation (e.g.~\cite{georgila_reinforcement_2011,lewis2017deal,cao2018emergent,eccles2019biases,hughes2020learning}). In particular, the environments considered involved only two agents, sidestepping the issue of coalition selection. Closer to our perspective is the work of~\cite{chalkiadakis_rl,matthews2012}, which propose a Bayesian reinforcement learning framework for team formation. However, they do not consider spatially extended environments and the computational cost of the Bayesian calculation is significant. 

{\bf Evaluation Games:} We evaluate our approach using two environments. In the {\it Propose-Accept game} (see Section~\ref{l_sect_propose_accept_env}), each agent has a weight, representing the proportion of the resources required to achieve a task. Agents have to form a team whose sum of weights exceeds 100\% of the required resources. A successful team obtains a unit of reward, that needs to be shared between the team members. Agents take turns making offers to a potential team of partners, proposing an allocation of the rewards. If the rest of the potential team members accept the proposal, the game terminates and the team members share the reward according to the proposal; if even one of the proposed team members declines, the game moves on to the next round and a proposer is chosen at random. In the {\it Team-Patches} game (see Section~\ref{l_sect_team_patches_env}, illustrated in Figure \ref{fig:env_team_patches_1}), agents inhibit a grid world with several colored rectangular areas called {\em patches}. Agents can move in the grid, and form a team with other agents standing in the same patch. Similarly to the Propose-Accept game, agents each have a weight, and a team is valid if the sum of the weights exceeds 100\% of the required total weight. Rather than making direct proposals, agents set a demanded share of the available reward to be a part of the team. The game terminates once there is a patch where the agents standing on the patch have a valid total weight, and once the total required share is at most a single unit. In other words, when the total demand of the team members exceeds the unit reward, agents are still not in agreement on how to share the reward so negotiation continues. 

We show that agents trained via independent reinforcement learning outperform hand-crafted bots based on game-theoretic principles. We analyze the reward distribution in the Propose-Accept game, showing a high correspondence with the Shapley value solution from cooperative game theory. We show that the slight deviation is not due to lack of neural network capacity by training a similar-sized {\em supervised} model to predict the Shapley value. 
We show that the correspondence with the Shapley value also holds in the spatial Team-Patches environment, and investigate how spatial perturbations influence agents' rewards.

\section{Preliminaries}
\subsection{Cooperative Games}
We provide definitions from cooperative game theory that our analysis uses (for a full review see~\cite{osborne1994course} and~\cite{chalkiadakis2011computational}). 
A (transferable-utility) {\em cooperative game} consists of a set $A$ of $n$ agents, and a {\em characteristic function} $v : P(A) \rightarrow \mathbb{R}$ mapping any subset $C \subseteq A$ of agents to a real value, reflecting the total utility that these agents can achieve when working together. We refer to a subset of agents $C \subseteq A$ as a {\em team} or {\em coalition}. A {\em simple cooperative game} has $v$ take values in $\{0 ,1\}$, where $v(C) = 1$ iff $C$ can achieve the task, modelling the situation where only certain subsets are viable teams. In such games we refer to a team $X$ with $v(X)=1$ as a {\em winning team}, and a team $Y$ with $v(Y)=0$ as a {\em losing team}. Given a winning team $C$ and an agent $a \in C$, we say that player $a$ is {\em pivotal} in a winning coalition $C$ if $v(C \setminus \{a\})=0$, i.e. $a$'s removal from $C$ turns it from winning to losing. An agent is pivotal in a permutation $\pi$ of agents if they are pivotal in the set of agents occurring before them in the permutation union themselves. Formally, let $S_{\pi}(i) = \{j | \pi(j) < \pi(i)\}$. Agent $i$ is pivotal in $\pi$ if they are pivotal for the set $S_{\pi}(i) \cup \{ i \}$. 
The {\bf Shapley value} characterizes fair agreements in cooperative games~\cite{shapley1953value}. It is the only solution concept fulfilling important fairness axioms, and is therefore an important quantity in cooperative games~\cite{dubey1975uniqueness,dubey1979mathematical,straffin1988shapley}. The Shapley value measures the proportion of all permutations in which an agent is pivotal, and is given by the vector $\phi(v) = (\phi_1(v), \phi_2(v), \ldots, \phi_n(v))$ where
\begin{equation}
\phi_i(v) = \frac{1}{n!} \sum_{\pi \in \Pi} [v(S_\pi(i) \cup \{i\}) - v(S_\pi(i))] \, .
\end{equation}

\subsubsection{Weighted voting games}
\label{l_sect_wvg}
Weighted voting games 
model situations where each agent is endowed with a certain amount of a resource and a task can only be achieved given a certain quantity of the resource (called the threshold or quota). Weighted voting games have been extensively used to analyze power in legislative bodies, where parties each have a certain number of parliament seats following an election, but a government can only be formed with at least a quota of half the total number of seats (some countries require an even higher quota for certain cases). 

A {\em weighted voting game} $[w_1, w_2, \ldots, w_n; q]$ is a simple cooperative game described by a vector of weights $(w_1, w_2, \ldots, w_n)$ and a quota (threshold) $q$. A coalition $C$ wins iff its total weight (the sum of the weight of its participants) meets or exceeds the quota. Formally $v(C) = 1$ iff $\sum_{i \in C} w_i \geq q$. By abuse of notation, we identify the game description with the characteristic function, writing $v = [w_1, w_2, \ldots, w_n; q]$. The Shapley value of weighted voting games has been used to analyze political power in legislative bodies, and for such settings it is referred to as the Shapley-Shubik power index~\cite{shapley1954method}.

\subsubsection{The Shapley Value as a Measure of Power in Cooperative Games}
\label{l_sect_appendix_power_in_politics_example}
We investigate the relation between agreements reached among RL agents and the Shapley value in a weighted voting game setting. To illustrate and motivate the importance of the Shapley value and other power indices, we now discuss several example domains.  Consider multiple providers such as travel agencies or airline carriers which can allow a person to travel between various destinations, and a client who is willing to pay a certain amount to get to a desired destination; while there is no direct flight to the client's destination, there are multiple routing options, using different carrier combinations. How should the carriers share the customer's payment? Similarly, consider a manufacturing scenario where multiple companies can provide subsets of components required to manufacture an end product, and where each company has only some of the components. How should they share the profit from selling the end product?  

Both the above scenarios can be captured as a cooperative game. 
\footnote{We provide a high level framework where RL agents can learn how to negotiate and form teams. Although our analysis focuses on a weighted voting game setting, our framework generalizes to many cooperative game settings, including ones such as those discussed in these scenarios.}
Solution concepts from cooperative game theory can analyze such domains, and make predictions regarding how agents are likely to share the joint gains. However, the most prominent example for applying cooperative game theory to analyze the negotiation power of agents originates from measuring political power in decision making bodies~\cite{straffin1988shapley,shapley1954method,felsenthal1998measurement}. We illustrate how power indices, and the Shapley value in particular, formalize power in a way that depends on the possible teams that can form, rather than the direct parameters of the game (such as the weights). 

Consider the formation of a coalition government following an election, where a set of political parties each obtained a certain number of parliament seats in an election, and where a quota of the majority of the seats is required to form a ruling government. If no single party has won the majority of the votes on its own, multiple parties would have to join so as to form a ruling coalition government. Parties in this setting would have to negotiate to form a ruling government. \footnote{The election outcomes in some countries, such as the USA or the UK, typically have a single party obtaining the majority of the votes. However, in many other countries typically no single party has an outright majority, requiring the formation of a coalition government.}

For intuition, consider a parliament of 100 seats, two big parties with  
49 seats each and a small party with 2 seats; a majority requires 50 seats (half of the total number of seats), so no party can form a government on its own. While each big party has far more seats than the small party, any team of two or more parties has the required majority of seats. Under such a framing of the domain, any team of at least two out the three parties is sufficient for completing the task of forming a government, and due to this symmetry between parties one may claim they all have equal power. In other words, what matters in making a party powerful is not its number of parliament seats, but rather the number of opportunities it has to form teams. Intuitively, one would say that the small party has a strong negotiation position. It could, for instance, demand control of a dis-proportionally high part of the budget (i.e. although it only has 2\% of the seats, it is likely to get control of a much larger share of a the budget). 

We note that for the above example, the Shapley value would give each party the same share. In fact, the Shapley value as a function does not even take the weights or threshold as input, but rather the characteristic function of the game. 
We say that agents $i,j$ are {\em equivalent} in a game with characteristic function $v$ if for any coalition $C$ that contains neither $i$ nor $j$ (i.e. any $C$ such that $i \notin C$ and $j \notin C$) we can add either $i$ or $j$ to the coalition and obtain the same value (i.e. $v(C \cup \{ i \}) = v(C \cup \{ j \})$). We note that in the above example, all agents are equivalent, and that the Shapley value allocates them the same share. This is not a coincidence --- one of the fairness axioms characterizing the Shapley value is that equivalent agents get the same share~\cite{shapley1953value,dubey1975uniqueness}. Indeed, the Shapley value is the only index fulfilling a small set of fairness axioms such as this one.

Two additional axioms that fully characterize the Shapley value, in addition to the above equivalence axiom, are that of giving null players no reward, and the additivity axiom. A null player is one which adds no value to any coalition, and the null player axioms states that null players would get no reward; the additivity axioms relates to the sum of multiple games, stating that the value allocated to any player in the sum-game would be the sum of the values in the individual composing games. For a more detailed discussion of such axioms, see textbooks and papers on the axiomatic characterization of the Shapley value~\cite{straffin1988shapley,chalkiadakis2011computational}.


\subsection{Multi-agent Reinforcement Learning}
\label{l_sect_marl}

An $n$-player {\em Markov game} specifies how the state of the an environment changes as the result of the joint actions of $n$ individuals. The game has a finite set of states $\cS$. The observation function $O: \cS \times \{1,\dots,n\} \rightarrow \bR^d$ specifies each player's $d$-dimensional view of the state space. We write $\cO^i = \{ o^i~|~s \in \cS, o^i = O(s,i)\}$ to denote the observation space of player $i$. From each state, players take actions from the set $\cA^1,\dots,\cA^n$ (one per player). The state changes as a result of the joint action $a^1,\dots,a^n \in \cA^1,\dots,\cA^n$, according to a stochastic transition function $\cT: \cS \times \cA^1 \times \cdots \times \cA^n \rightarrow \Delta(\cS)$, where $\Delta(\cS)$ denotes the set of 
probability distributions over $\cS$. Each player receives an individual reward defined as $r^i: \cS \times \cA^1 \times \cdots \times \cA^n \rightarrow \bR$ for player $i$.

In {\em multi-agent reinforcement learning} each agent independently learns, through its own experience,
a behavior policy $\pi^i : \cO^i \rightarrow \Delta(\cA^i)$ (denoted $\pi(a^i|o^i)$) based on its observation $o^i$ and reward $r^i$. 
Each agent's goal is to maximize a long term $\gamma$-discounted payoff~\cite{SuttonBarto18}. Agents are independent, so the learning is {\it decentralized}~\cite{Bernstein02}.

\section{Game Setting} 
\label{l_game_setting}

\subsection{Overview}
\label{l_sect_overview}
We propose a method for training agents to negotiate team formation under a diverse set of negotiation protocols. In this setting, many different combinations of agents can successfully perform a task. This is captured by an {\em underlying cooperative game}, given by a characteristic function $v : P(A) \rightarrow \{0, 1 \}$ (which maps viable teams $C \subseteq A$ to the value 1, and non-viable teams to 0). 
When a viable team is formed, it obtains a reward $r$, to be shared between the team's individual members. The outcome of the team formation process is an agreement between agents of a viable team regarding how they share the reward $r$ (with each agent trying to maximize its share). Cooperative game theory can characterize how agents would share the joint reward, by applying solution concepts such as the Shapley value. However, cooperative game theory abstracts away the mechanism by which agents negotiate. To allow agents to interact, form teams and reach an agreement regarding sharing the reward, one must use a {\em negotiation protocol}, forming an environment with specific rules governing how agents interact; this environment consists of the actions agents can take, and their semantics, determining which team is formed, and how the reward is shared.

\begin{figure}
  \centering
  \includegraphics[width=\textwidth, height=0.15\textheight]{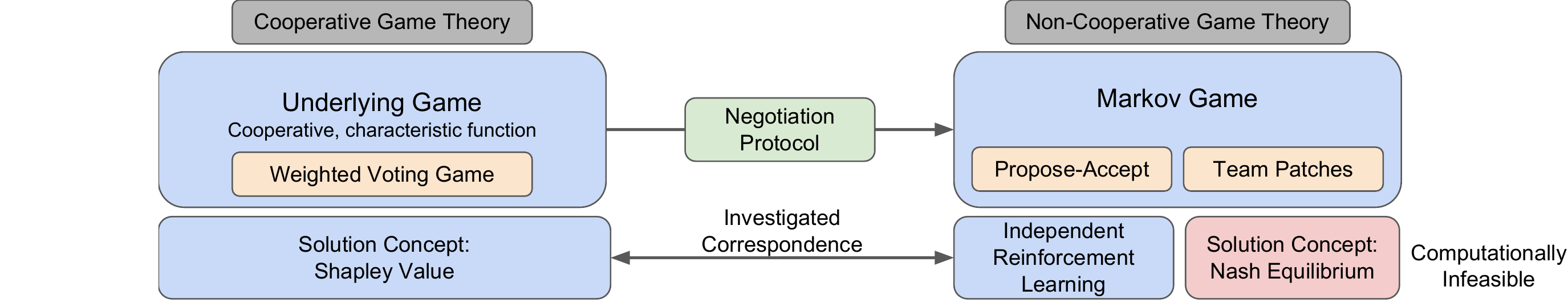}
  \caption{An overview of our methodology. We examine a team formation task defined by an underlying cooperative game. Applying different negotiation protocols to the same underlying task generates different environments (Markov games). Instead of hand-crafting negotiation bots to each such environment, we train independent RL negotiation agents. We compare the agreements RL agents arrive at to game theoretic solutions to the underlying cooperative game. }
  \label{fig:overview}
\end{figure}

We examine two simple negotiation protocols, a non-spatial environment where agents take turns making offers and accepting or declining them, and a spatial environment where agents control their demanded share of the reward in a grid-world setting. Overlaying the underlying cooperative game with a specific negotiation protocol yields a {\em negotiation environment}; this is a Markov game, which 
may be analyzed by non-cooperative game theory, identifying the equilibrium strategies of agents. However, solving even the simplest such games may be computationally infeasible: even the restricted case of an unrepeated two-player general-sum game is computationally hard to solve, being PPAD-complete~\cite{ChenDeng06} (we discuss this issue further in Section~\ref{l_sect_conclusions}). 
Instead, we propose training independent RL agents in the negotiation environment. While our methodology can be applied to any cooperative game, our experiments are based on a weighted voting game as the underlying cooperative game. We examine the relation between the negotiation outcome that RL agents arrive at and the cooperative game theoretic solution, as is illustrated in Figure~\ref{fig:overview}.

\subsection{Negotiation Environments}
\label{l_sect_nego_env}

Our negotiation environments use a weighted voting game $[w_1, \ldots, w_n; q]$ as the underlying game, offering a total reward $r \in \mathbb{N}$. Each agent $i$ is assigned a weight $w_i \in \mathbb{R}$ at the beginning of each episode, and the goal of the agents is to construct a team $C$ whose total weight $\sum_{i \in C} w_i$ exceeds a fixed quota $q$. If a successful team forms, the {\em entire} team is assigned the fixed reward $r$. The agents comprising the team must decide how to allocate this reward amongst themselves by agreeing on shares $\{ r_i \}_{i \in C}$ such that $\sum_{i \in C} r_i = r$, with $r_i \in \mathbb{N}_0$. We say a team $C \subseteq A$ is {\em viable} if $\sum_{i \in C} w_i \geq q$. Though there are many viable teams, not all teams are viable (depending on the weights). Only one viable team is chosen, and non-members all get a reward of $0$. In such settings, agents face opportunities to reach agreements for forming teams, along with their share of the gains. If they do not agree, they stand the risk that some other team would form without them (in which case they get no reward); on the other hand, if they agree to a low share of the reward, they miss out on the opportunity to reach an agreement with others giving them a potentially higher share.   


Our two negotiation environments have an identical underlying game,
but employ different protocols for negotiating over the formed team and reward allocation. The non-spatial environment uses direct proposals regarding formed teams and reward allocations, 
while our spatial environment has a similar interpretation to that of recent work applying multi-agent deep reinforcement learning to non-cooperative game theory settings like spatially extended seqential social dilemmas (e.g.~\cite{leibo2017multi,perolat2017multi,hughes2018inequity}). 

\subsubsection{Propose-Accept (Non-Spatial)}
\label{l_sect_propose_accept_env}
In the Propose-Accept environment, agents take turns in proposing an agreement and accepting or declining such proposals. The underlying game, consisting of the weights and threshold, is public knowledge, so all agents are aware of $v=[w_1, \ldots, w_n; q]$ chosen for the episode. Each turn within the episode, an agent is chosen uniformly at random to be the {\em proposer}. The proposer chooses a viable team and an integer allocation of the total reward between the agents. The set of actions for the proposer agent consists of all n-tuples $(r_1, r_2, \ldots, r_n)$, where $r_i \in \mathbb{N}_0$ and $\sum_{i=1}^n r_i = r$. By convention, the selected team $C$ consists of the agents with non-zero reward under this allocation; that is to say $C=\{ i | r_i > 0 \}$. 

We refer to the agents in the team chosen by the proposer as the {\em proposees}. Once the proposer has made a proposal, every proposee has to either {\em accept} or {\em decline} the proposal. If all proposees accept, the episode terminates, with the rewards being the ones in the proposed allocation; if one or more of the proposees decline, the entire proposal is declined. When a proposal is declined, with a fixed probability $p$ a new round begins (with another proposer chosen uniformly at random), and with probability $1-p$ the game terminates with a reward of zero for all agents, in which case we say the episode terminated with agents {\em failing to reach an agreement}. 

All proposals consist of a viable team and an allocation $(r_1, \ldots, r_n)$ such that $\sum_{i=1}^n r_i = r$, so the total reward all the agents obtain in each episode, is either exactly $r$ (when some proposal is accepted), or exactly zero (when agents fail to reach an agreement). 
Interactions in the Propose-Accept environment can be viewed as a {\em non-cooperative} game, however solving this game for the equilibrium behavior is intractable (see Section~\ref{l_sect_conclusions} for details).

\subsubsection{Team Patches (Spatial)}
\label{l_sect_team_patches_env}
\begin{figure}[htbp]
\centering
  \includegraphics[width=0.60\textwidth]{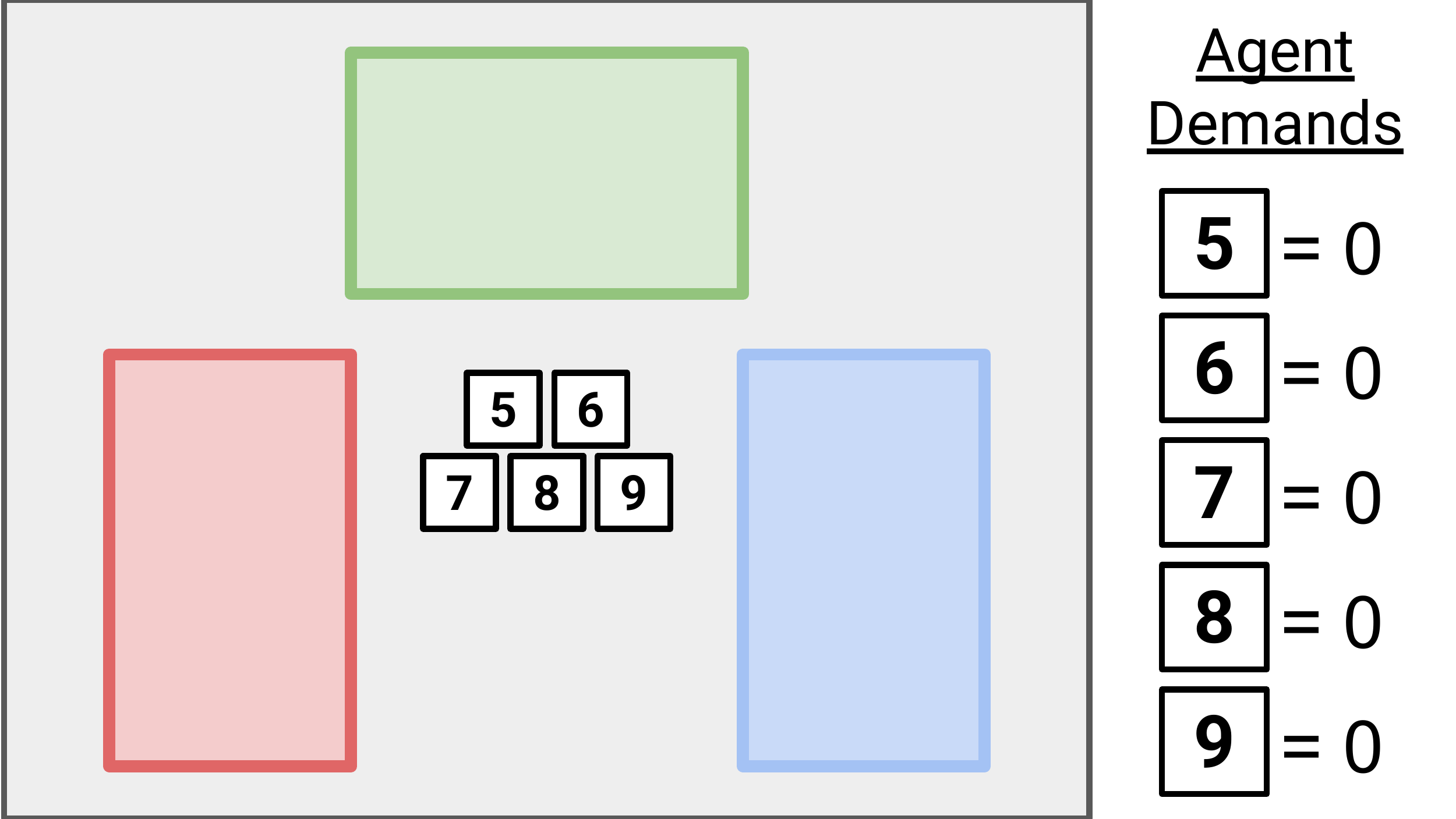}
  \caption{The Team Patches Environment, beginning of an episode.}
  \label{fig:env_team_patches_1}
\end{figure}

We construct a spatial negotiation environment, based on the same underlying weighted voting game as in the Propose-Accept environment, called the Team Patches environment (shown in Figure \ref{fig:env_team_patches_1}). This is a $15\times15$ grid-world that agent can navigate, including several colored rectangular areas called {\em patches}. Agents can form a team with other agents occupying the same patch, demanding a share of the available reward to be a part of the team. Similarly to Propose-Accept, the underlying game (i.e. the agent weights and the quota) are fully observable by all agents. Additionally, as the environment is spatial, agents also observe their spatial surrounding (including the locations of patches and of other agents).

\begin{figure}[htbp]
\centering
  \includegraphics[width=0.60\textwidth]{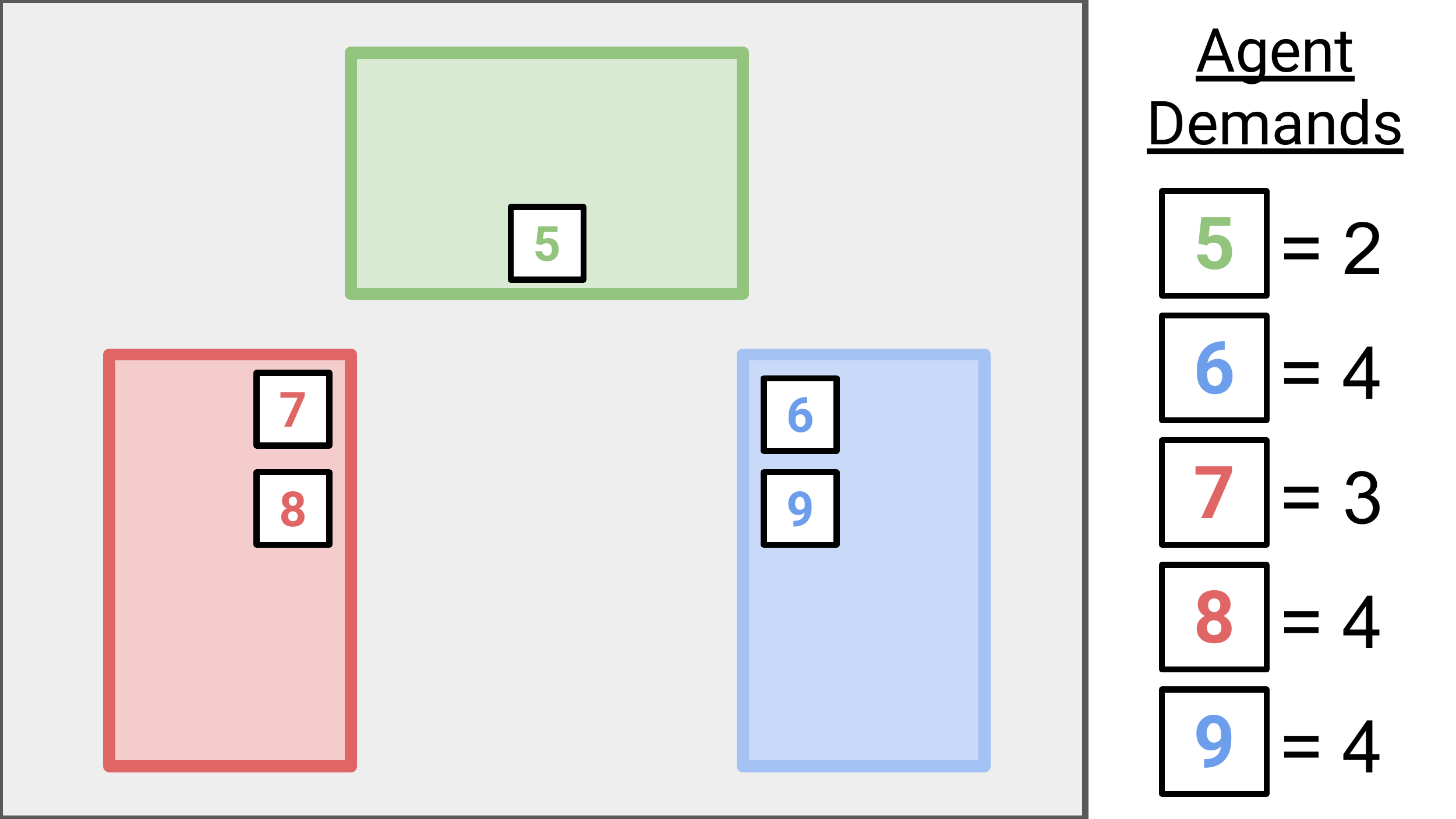}
  \caption{The Team Patches Environment, termination of an episode.}
  \label{fig:env_team_patches_2}
\end{figure}

At the start of an episode, agents are randomly initialized in the center of the map. Each step, they can move around the environment (forwards, backwards, left, right), rotate (left, right), and set their {\em demand} ($d_{i} \in \{ 1, 2 \ldots, r \}$), indicating the minimal share of the reward they are willing to take to join the team. To form teams, agents must move into patches, and we refer to all agents occupying the same patch as the patch's team. The team $T^j$ in patch $j$ is {\em viable} if the total weight of the agents in $T^j$ is greater than the threshold ($\sum_{i \in T^j} w_{i} \geq q$). The demands of the patch's team are {\em valid} if the total demanded is less than the total available ($\sum_{i \in T^j} d_i \leq r$). An agreement is reached once there is a patch whose team is both viable and with a valid reward allocation. An episode terminates when an agreement is reached on one of the patches $j$, giving each agent $i$ in the team $T^j$ their demanded reward ($d_{i}$). Unlike the Propose-Accept environment, the agents don't necessarily use up all the reward available, since $\sum_i d_i < r$ is allowed. If the agents fail to reach an agreement after $100$ steps, the episode is terminated, and all agents receive $0$ reward. An example for an episode termination is given in Figure~\ref{fig:env_team_patches_2} and discussed below. 

Figure~\ref{fig:env_team_patches_1} shows the Team Patches environment for a game with threshold $q=15$ and a total reward of $r=7$. Agents are represented by the squares containing their weights. In this example, we have $n=5$ agents, with the weights $v=[5,6,7,8,9]$, and three patches where agents can form teams (red (left), green (top), and blue (right)). As shown in the figure, at the start of an episode, the agents randomly begin in the center of the map with no assigned team or demands. 

Figure~\ref{fig:env_team_patches_2} shows an example configuration for the same game as in Figure~\ref{fig:env_team_patches_1}, but at the end of an episode, where the agents have moved inside the grid-world to each of the patches. In this example, the agents in red (7 and 8) form a viable team with a valid reward allocation as their weights are above the required threshold ($7+8 \geq 15$) and their demands are equal to the availability ($3+4 \leq 7$). The team in green is not viable as the total weight is not sufficient ($5 \ngeq 15$), and the blue team has an invalid reward allocation as their demands are higher than the availability ($4+4 \nleq 7$). Agents 7 and 8 receive 3 and 4 reward respectively, and the rest of the agents receive 0 reward.

\subsection{Learning Agents}

We apply independent multi-agent reinforcement learning to train agents to negotiate in the Propose-Accept and Team-Patches environments. The required inputs to begin the simulations are the parameters of the underlying weighted voting games, which include the number of agents $n$, the agents weights $w_1,\ldots,w_n$ and the threshold $q$ for a successful coalition (see Section~\ref{l_sect_wvg}). We further require the parameters of the specific environment (Propose-Accept or Team Patches), such as the locations of the agents patches (for Team Patches) and the number of reward units $r$ (for Propose-Accept). Finally, the simulation depends on the hyperparameters of the learners (such as the learning rate $\alpha$). 

We briefly describe the computational experience of training the agents. Our simulation is based on independent multi-agent reinforcement learning (see various textbooks and surveys providing a detailed discussion of such methods~\cite{Littman94markovgames,shoham2003multi}). As discussed in Section~\ref{l_sect_marl}, we maintain multiple independent reinforcement learning agents, each capturing a policy mapping observations regarding the state of the environment to an action to take. In each episode, each of the agents chooses actions based on their current policy: in every state, agents select actions based on partial observations of the true world state, each receiving their individual reward. This results in experience in the form of a trajectory representing a single episode (game) in the environment, consisting of the sequence of states the agent has observed, the action taken in that state, and the reward the agent achieved. Through their individual experiences interacting with one another in the environment, agents learn an appropriate behavior policy, using a reinforcement learning algorithm, which updates an agent's policy given the experience gathered (the trajectory). 

For Propose-Accept, each agent independently learns a policy using SARSA($\lambda$)~\cite{rummery94} with $\lambda = 0.1$. We use a function approximator to learn the $Q$ function, 
reducing the intractable state space to manageable number of features, and apply a neural network as our function approximator. 
The weights of the network are trained with the 
Adam optimizer~\cite{kingma2014adam} to minimize the online temporal difference error. The function approximator is a multi-layer perceptron with $3$ hidden layers, each of size $64$.

For Team Patches, which is a spatial environment, we use an advantage actor-critic algorithm~\cite{mnih2016} with the V-trace correction~\cite{espeholt2018}, 
employing a convolutional neural network followed by a multi-layer perceptron, and learning from 16 parallel copies of the environment with an on-policy algorithm. 
The neural network uses a convolutional layer with $6$ channels, kernel size $3$ and stride $1$, followed by a multi-layer perspective with $2$ hidden layers of size $32$. The policy and value function heads are linear layers. Our 
training 
setup is analogous to~\cite{jaderberg2018human}, but 
using no evolution. 

In some of our experiments we compare our RL agents with hand-crafted bots. We emphasize that our RL agents are trained with one another~\cite{Littman94markovgames,shoham2003multi}, and we only compare them with hand-crafted bots following training, at which point we freeze their parameters, stopping them from learning. As a result, they do {\it not} learn to exploit any weaknesses of the bots.

\section{Experiments}
We now present our four key experimental results. First, we demonstrate that training reinforcement learning agents to negotiate leads to sensible strategies, outperforming hand-crafted bots designed for such scenarios. Further, we show that our agents generate negotiation outcomes consistent with fair solutions predicted by cooperative game theory, in both non-spatial and spatially extended environments. We then investigate how spatial structure can influence negotiation outcomes. Finally, we test whether the deviations from the game theoretic predictions are due to the representational capacity of our agents or their learning dynamics.

\subsection{Experimental Setup}
Our experiments are based on a distribution $D$ over underlying weighted voting games. We sample games from this distribution, and refer to them as the experiment's {\em boards}; each board consists of its weights and threshold $(v^i = [w^i_1, \ldots, w^i_n; q^i])$. Each such board has $n=5$ agents, a threshold of $q = 15$, and weights sampled from a Gaussian distribution $w_i \sim \mathcal{N}(6,1)$, and exclude boards where all players have identical power. 

We partition the boards sampled from the distribution to a {\em train set} and a {\em test set}. Agents are trained for $500,000$ games using train set boards and are evaluated on the test set, requiring them to learn a general strategy for negotiating under different weights, 
and to be have to generalize to negotiation situations they have not encountered during training. We use a train set of 150 unique boards and a test set of $k=50$ unique boards. 

\subsubsection{On the Choice of the Weight Distribution and Convergence}
\label{l_sect_choice_weight_dist_convergence}
We use a Gaussian distribution as Weighted Voting Games with Gaussian weight models have been previously investigated in the literature on computational game theory~\cite{chalkiadakis2011computational,bachrach2013reliability}. We note that the methods we propose here work for any other weight distribution as well. However, the relative differences in player power, as measured by the Shapley value, may be quite different in such games. For instance, in certain distribution we are more likely to encounter outliers in terms of weights and Shapley values. Some recent work has examined the influence of outliers and output constraints on the nonlinear and real systems under non-Gaussian noise models~\cite{filipovic2011robust,stojanovic2016optimal,stojanovic2016joint}.

In general, independent multi-agent reinforcement learning may not converge to a Nash equilibrium in our game. While there are some known convergence results for such procedures in two-player zero-sum games~\cite{shoham2003multi,bu2008comprehensive,lanctot2017unified}, our game has more than two players and is not zero-sum (for example, in some cases all agents get no reward and in other cases some agents have positive utility). As a result, agents may cycle through policies during training, at least for some weight configurations, in a way that depends on the weight configuration. 
In addition to the above methods for dealing with non-Gaussian noise, one could consider alternative agent learning algorithms, such as Fictitious Play~\cite{brown1951iterative,hofbauer2002global,shamma2005dynamic}, variants of no-regret learning~\cite{bowling2005convergence} or alternative learning dynamics~\cite{conitzer2007awesome}. 

\subsection{Comparison with Hand-Crafted Bots}
\label{l_sect_hand_crafted_bot_comparison_short}
We compare the negotiation performance of RL agents trained in our framework with hand-crafted bots. While hand-crafted bots can form excellent negotiators, they can only be used for a specific negotiation protocol, and some of them require human data to construct. 
We note that although negotiation protocols differ across environments, the essence of the decisions agents face is similar. When agents fail to reach an agreement they obtain no reward, so agents who almost never reach an agreement would be bad negotiators. On the other hand, reaching agreement easily by itself does not make an agent a strong negotiator. For instance, an agent who accepts any offer is likely to reach agreement easily but perform poorly, as they would take even very low offers (even when they have a high relative weight and thus a strong negotiation position). 

Although the essence of the task is similar across environments, the observations, action spaces and semantics of interaction differ considerably across negotiation protocols. The key advantage of using RL to train negotiation agents is not having to hand-craft rules relating to the specifics of the negotiation environment. Indeed, as hand-crafting a bot is a time consuming process, we focused on the simpler propose-accept environment; creating a hand-crafted bot for the team patches environment is more difficult, as there are many more decisions to be made (how to find a good patch, which share to demand, how to respond if the total reward demanded is too high, etc.)

For simplicity, we examine the Propose-Accept protocol where a negotiation bot faces two kinds of decisions: (1) what offer to put in as a proposer, and (2) which offers to accept as a proposee. Our simplest baseline is a {\em random bot}. As a proposer, it selects an allowed proposal uniformly at random from the set of all such proposals. As a proposee it accepts with probability $\frac{1}{2}$ and rejects with probability $\frac{1}{2}$. 
A more sensible baseline is the {\em weight-proportional bot}, which is an adaptation of previously proposed negotiation agents to the propose-accept protocol~\cite{lin2008negotiating,lin2010can,baarslag2012first,mash2017form} (the propose-accept protocol itself bears some similarity to the negotiation protocols used in such earlier work on negotiation agents). An even stronger baseline is the {\em Shapley-proportional bot}, which is similar to the weight-proportional bot, but computes the Shapley values of all the agents (which is computationally tractable only for games with few agents).

As a proposer, the weight-proportional bot randomly chooses a viable team $C$ which contains itself from the set of all such teams. It then proposes an allocation of the reward that is proportional to each the team agents' weights. Formally, given a board $[(w_1, \ldots, w_n); q]$, we denote the total weight of a team $C$ by $w(C) = \sum_{i \in C} w_i$. For the team $C$ it chose, the bot uses the target allocation $p_i = \frac{w_i}{w(C)} \cdot r$ where $r$ is the fixed total reward. \footnote{Our framework's action space includes only integral reward allocations, so the bot chooses the integral reward allocation that minimizes the $L_1$ distance to the target allocation.} 

As a proposee, the weight-proportional bot computes its share of the same target allocation $p_i = \frac{w_i}{w(C)} \cdot r$ in the proposed team $C$, and compares it with the amount $r_i$ offered to it in the proposal. We denote by $g_i = r_i - p_i$ the amount by which the offer $r_i$ exceeds $p_i$, the bot's expectations under the weight-proportional allocation. A high positive amount $g_i$ indicates that the bot is offered much more than it believes it deserves according to the weight-proportional allocation, while a negative amount indicates it is offered less than it thinks it deserves (when offered exactly the weight-proportional share we have $g_i = 0$). The probability of the bot accepting the offer is $\sigma(c \cdot g_i)$ where $\sigma$ denotes the logistic function $\sigma(x)=\frac{1}{1+e^{-x}}$, and where $c=5$ is a constant correcting the fact that $g_i$ only ranges between $-1$ and $+1$ rather than between $- \infty$ and $+ \infty$. Thus the bot accepts a ``fair'' offer (according to the weight-proportional target) with probability of $\frac{1}{2}$, and the probability convexly increases as $g_i$ increases (and decreases as $g_i$ decreases).

The more sophisticated {\em Shapley-proportional} bot follows the same design as the weight-proportional bot, except it sets the target allocation to be the one proportional to the Shapley values rather than the weights, i.e. $p_i = r \phi_i / (\sum_{i \in C} \phi_i)$ (where $\phi_i$ denotes the Shapley value of agent $i$). We note that as computing the Shapley values is an NP-hard problem~\cite{elkind2009computational} this method is tractable for 5 agents, but does not scale to games with many agents.

We compare the performance of the bots and RL trained agents by creating many co-trained groups of agents. 
We examine each of the evaluation boards (sampled from the distribution of Gaussian weigh boards). For each evaluation board, we create $t=200$ {\em pairs} of agent groups, where in each pair we have one group of $n=5$ independent RL agents (the {\em all-RL group}), and one set consisting of 1 bot and 4 independent RL agents (the {\em single-bot group}). Each group is co-trained for $m=500,000$ episodes, where during training each episode uses a different board from the train set. During the evaluation step, we let each agent group play $5,000$ games on each evaluation board in the test set. We repeat the same analysis, each time allowing the bot to play a different weight in the board (i.e. letting the bot be the agent in the first position, in the second position and so on).

We investigate the performance of the RL agent from the all-RL group and the bot from the single-bot group, when playing the same weight on the same evaluation board. We average the fraction of the total fixed reward achieved by the RL-agent and bot over the $t=200$ group pairs and over the $5,000$ episodes played with each evaluation board, and examine the difference $d$ between the amount won by the RL-agent and the amount won by the bot. A positive value for $d$ indicates the RL-agent has a better performance than the bot, and a negative number shows a higher performance for the bot. 

On average (across boards and episodes played on each evaluation board), the RL-agent outperforms the weight-proportional bot, achieving $0.025 \cdot r$ more of the total reward. In the all-RL group, each agent makes on average $0.203 \cdot r$, whereas in the one-bot group the bot makes on average $0.178 \cdot r$. In other words, the RL-agent obtains 10\% more reward than the bot. We performed a Mann-Whitney-Wilcoxon test, which shows the difference is significant at the $p<0.005$ level. 
\footnote{The Mann-Whitney-Wilcoxon test is a non-parametric test similar to Student's T-test, but resistant to deviations from a Gaussian distribution.}

The results are similar for the Shapley-proportional bot, though the performance gap is smaller, with the Shapley-proportional bot making on average $0.185 \cdot r$ (the RL agent outperforms the Shapley-proportional bot at the same statistical significance level as for the weight-proportional bot). Unsurprisingly, performing a similar analysis for the random bot, that selects valid actions uniformly at random, we get even more significant results ($p < 0.001$). 

We have performed an additional experiment, examining the performance of RL agents against ``out-of-distribution'' bots. In this experiment, we trained the RL-agents in the one-bot group with a weight-proportional bot, but used a Shapley-proportional bot during their evaluation. In other words, the RL agents adapt their policy against one type of bot, but are measured against another type of bot. This hindrance sightly lowers the average reward share of RL agents, allowing the bot to gain a slightly larger reward share of $0.188 \cdot r$. However, even in this case the RL-agents outperform the bot ($p < 0.005$).    

These results indicate that RL agents can outperform {\em some simple heuristic} hand-crafted bots in team formation settings. Our results certainly do not mean that it is impossible to create well-crafted bots, tailored to a specific negotiation protocol, that would out-negotiate RL agents. For instance, even for the weight-proportional and Shapley-proportional bots we can tune the distribution parameter $c$ discussed above, and possibly improve the negotiation performance. We view our analysis as an indication that RL-agents make at least somewhat ``reasonable'' or ``sensible'' negotiation decisions. The key advantage of the methodology we propose is its generality and ability to tackle new negotiation protocols, without relying on hand-crafted solutions. In other words, our technique offers a way of automatically constructing at least reasonable negotiators, without requiring fitting a bot to a given negotiation protocol. 

\subsubsection{Robustness to Agent Architecture and Hyperparameter Choices}

We note that training multiple agents through independent reinforcement learning results in a complex dynamic system, which can be difficult to model. In particular, all agents change their policy across time, so from the standpoint of each individual agent, they are facing a non-stationary environment. As a result, our results could be sensitive to the choices of the initial parameter configuration or the hyperparameters of the learning procedure, such as the use learning rate, the time discount factor $\lambda$ or the optimization method used. The above results contrasting RL agents and bots are robust to various changes to agent architecture choices and hyperparameter settings. RL agents outperform the bots, at the same statistical significance, even when perturbing the learning rate, hidden layer sizes, the allotted number of training rounds and the $\lambda$ parameters by $\pm 25\%$. \footnote{We emphasize that strong hand-crafted bots may well achieve a better performance for a {\em specific} negotiation protocol, at the cost of significant engineering to build the bot. The experiment is designed to show the RL agents do a ``reasonably good'' job at negotiation.}

Our results above are given for initial architecture and hyper-parameter settings that we have tried, rather for an architecture or hyperparameters tuned to identify the best negotiation agents. To tune parameters so as to maximize the negotiation performance of the resulting agents, one may either conduct extensive sweeps, or use various optimization methods for the neural architecture~\cite{zoph2016neural,elsken2018neural} or hyperparameters~\cite{bengio2000gradient,domhan2015speeding,maclaurin2015gradient}. Similarly, meta-heuristic and nature inspired optimization algorithms can also be used to tune parameters in order to improve agent performance~\cite{yang2010nature,fister2013brief,nedic2014optimal,stojanovic2016nature,prvsic2017nature}

\subsection{Consistency with Game-Theoretic Predictions} 
\label{l_sect_experiment_shapley}
We now discuss another experiment, investigating whether our RL trained agents negotiate in a way that is consistent with solution concepts from cooperative game theory, focusing on the Shapley value. As the Shapley value was proposed as a ``fair'' solution, one can view this as studying whether independent-RL agents are likely share the joint gains from cooperation in a fair manner. 

\begin{figure}[htbp]
\centering
    \begin{subfigure}{1.0\textwidth}
        \centering
        \includegraphics[width=0.49\textwidth]{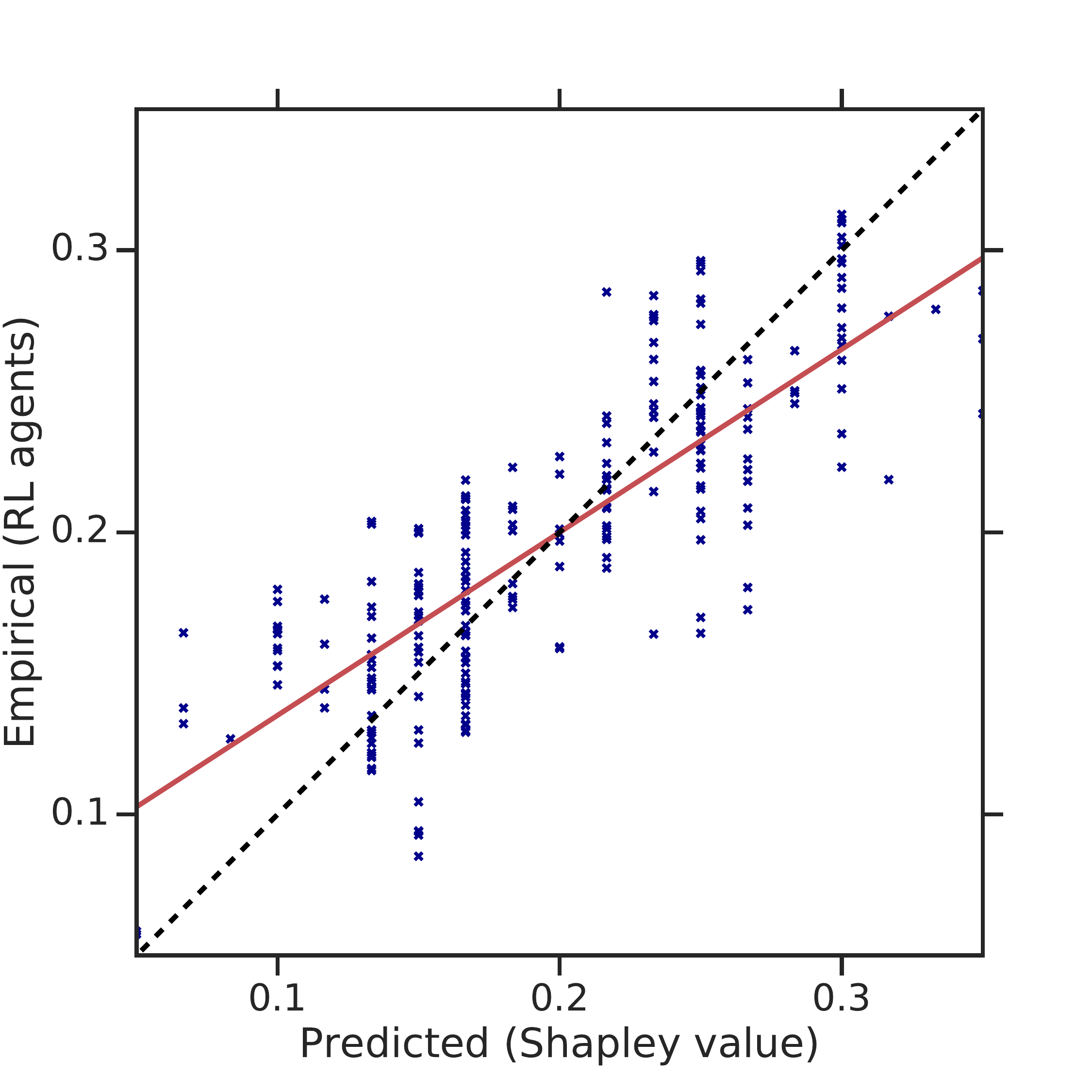}
        \includegraphics[width=0.49\textwidth]{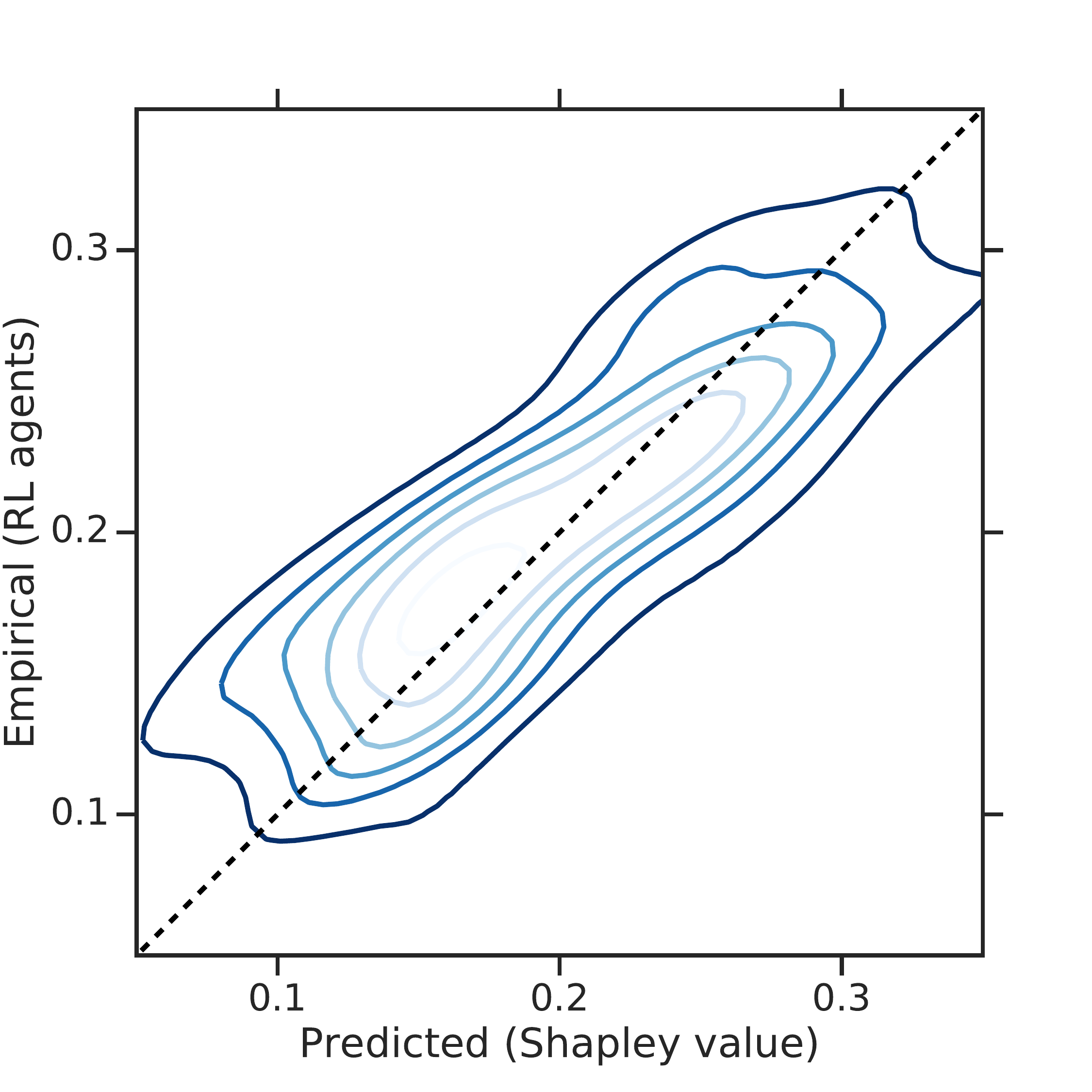}  
    \end{subfigure}
  \caption{Correspondence between Shapley values and empirical reward shares: Propose-Accept.}
  \label{fig:propose_accept_shapley}
\end{figure}

\begin{figure}[htbp]
\centering
    \begin{subfigure}{1.0\textwidth}
        \centering
        \includegraphics[width=0.49\textwidth]{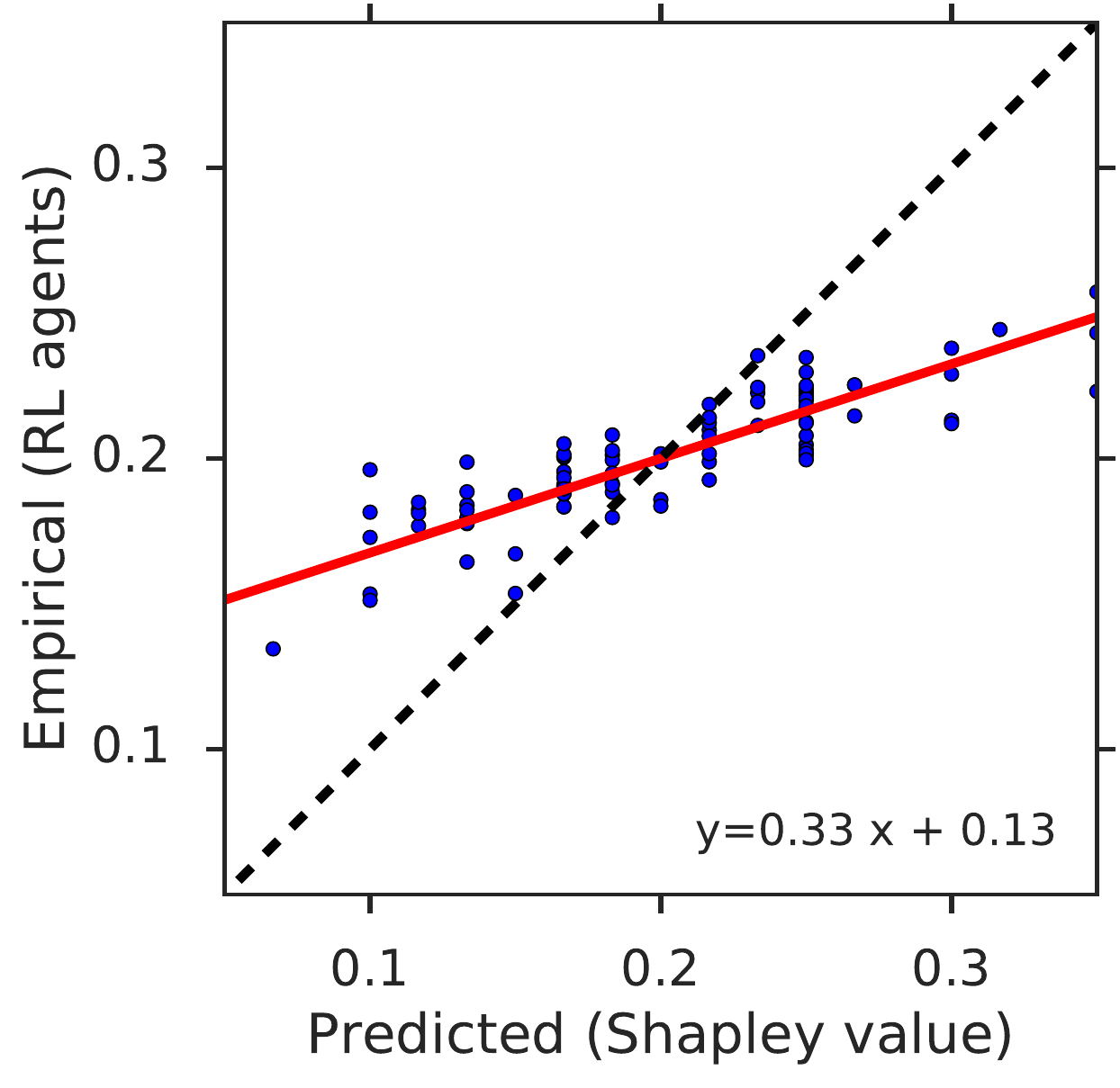}
        \includegraphics[width=0.49\textwidth]{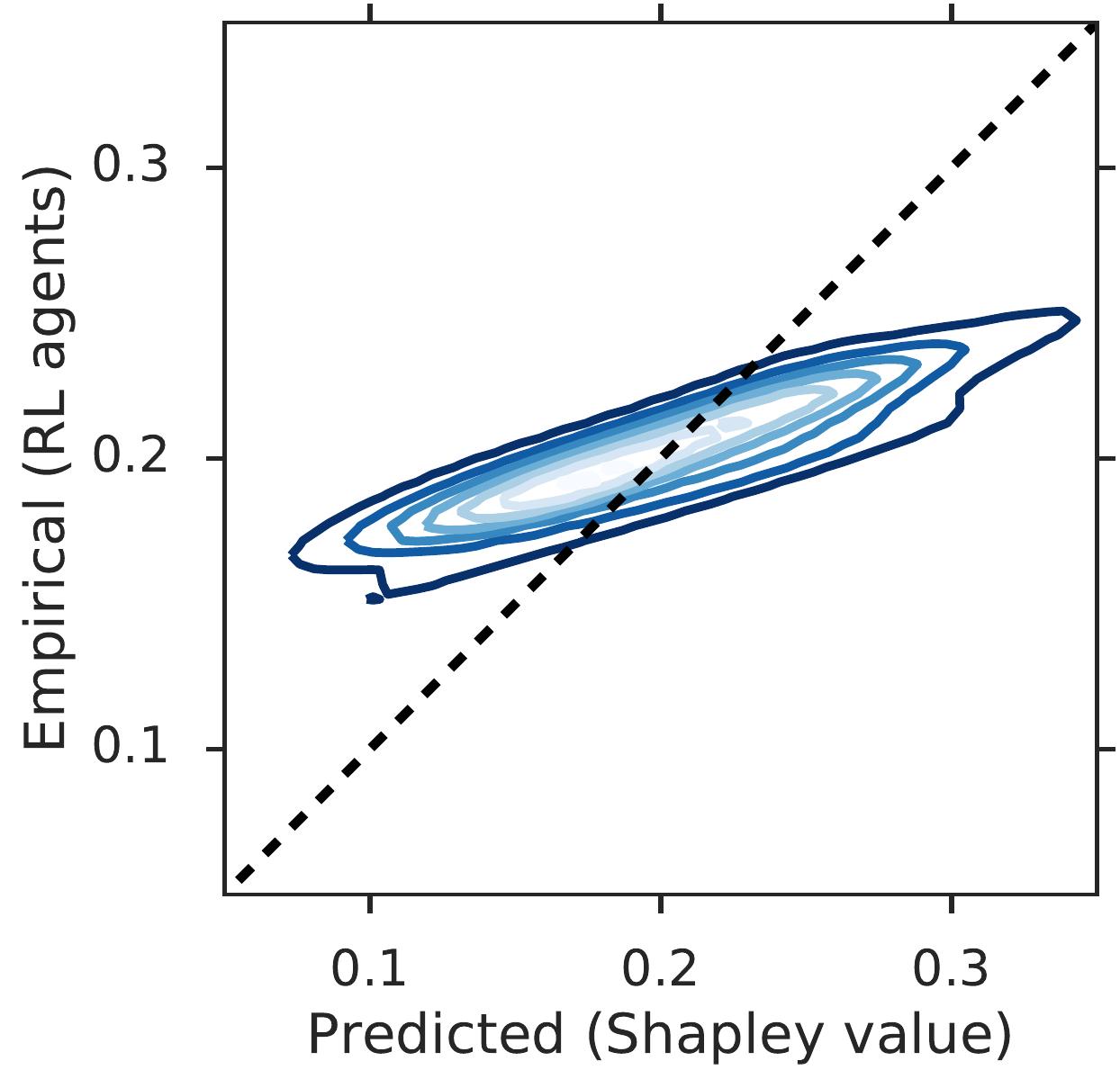}  
    \end{subfigure}
  \caption{Correspondence between Shapley values and empirical reward shares: Team Patches.}
  \label{fig:team_patches_shapley}
\end{figure}

To investigate the negotiation policies of our RL agents, we train and evaluate them on our two environments using the same set of $k=20$ boards. We denote the average amounts won by weight $i$ on board $j$ as $s^j_i$, and refer to these as {\em empirical agent rewards}.\footnote{For Propose-Accept, we average over 200 independent runs of 5000 episodes each, and for Team Patches we average over 100 independent runs of 1000 episodes each.} Given the game $v^j$ we can compute the Shapley values, denoted as $\phi(v^j) = (\phi^j_1, \ldots, \phi^j_n)$. We view $\phi^j_i$ as the game theoretic prediction of the fair share the agent with weight $w^j_i$ should get in a negotiation in the game $v^j$. We apply the analysis to each of our $k=20$ boards and obtain $k \cdot n$ pairs $\{ (\phi^j_i, s^j_i) \}_{i \in [n], j \in [k]}$ of Shapley value predictions and empirical agent rewards. Figure~\ref{fig:propose_accept_shapley} and Figure~\ref{fig:team_patches_shapley} show scatter plots and density estimation contours for these pairs. 





Figure~\ref{fig:propose_accept_shapley} shows the correspondence between agent Shapley values and the empirical share of the rewards these agents achieved in the Propose-Accept environment (i.e. the pairs $\{ (\phi^j_i, s^j_i) \}_{i \in [n], j \in [k]}$). On the left side of the figure we show a scatter plot, where the x-axis location of each point relates to the fair share prediction using the Shapley value, and the y-axis is the empirical reward share of the RL co-trained agents. We include $y=x$ (black dashed) to show the reward our agents would receive if they exactly matched the predictions, as well as a trend line (in red). On the right side of the figure we show the corresponding density-estimation contours, highlighting where most of the probability mass lies. 

Figure~\ref{fig:team_patches_shapley} shows the correspondence between agent Shapley values and the empirical share of the rewards these agents in the Team-Patches environment (equivalent to Figure~\ref{fig:propose_accept_shapley} which analyzed the Propose-Accept environment). Again, the left hand side shows a scatter plot, and the right hand side shows the corresponding density-estimation contours. 


Our results, depicted in Figure~\ref{fig:propose_accept_shapley} and Figure~\ref{fig:team_patches_shapley}, indicate a strong correlation between the Shapley value based predictions and the empirical gains of our independent RL trained negotiator agents. We see that the majority of the mass occurs in close proximity to the line $y=x$, indicating a good fit between the game theoretic prediction and the point to which our RL agents converge. RL agents exhibit a stronger deviation from the game theoretic ``fair'' reward on the boards where there are players with significantly more power or significantly less power than others, i.e. boards where there is a high inequality in Shapley values. 
In these boards, ``strong'' players (according to the Shapley value) do get a higher share than ``weak'' players, but not to the extent predicted; for these rare boards, the empirical allocation of reward achieved by the RL-agents is more equal than predicted by the Shapley value. Section~\ref{board_weight_variance_and_shapley_correspondence} examines how weight and power inequality affect the correspondence between the Shapley value and the outcomes achieved by RL agents, and Section~\ref{l_sect_shapley_distil} investigates potential reasons for the deviation from the Shapley value based predictions.

Our results show independent RL negotiation agents empirically achieve rewards that are consistent with the outcomes predicted by the Shapley value from cooperative game theory. Notably, this consistency extends to our spatial negotiation environment even with its different negotiation protocol and spatial interactions.


\subsubsection{The Impact of Weight and Power Inequality on the Correspondence with the Shapley Value}
\label{board_weight_variance_and_shapley_correspondence}

Section~\ref{l_sect_experiment_shapley} discussed the correspondence between the outcomes achieved by RL agents and the Shapley value from cooperative game theory, indicating that deviations occur mostly in boards where some agents have a particularly weak or particularly strong negotiation position, as measured by the Shapley value. The board distribution $D$ that we have there used ruled-out boards where all agents have identical weights (resulting in identical power). To better understand the impact of weight inequality on the agents' negotiation position, we now extend that analysis, showing that having a lower weight variance in the board distribution leads to a higher correspondence between the Shapley value and the empirical gains of RL agents. 

\begin{figure}[htbp]
\centering
    \begin{subfigure}{1.0\textwidth}
        \centering
        \includegraphics[width=0.49\textwidth]{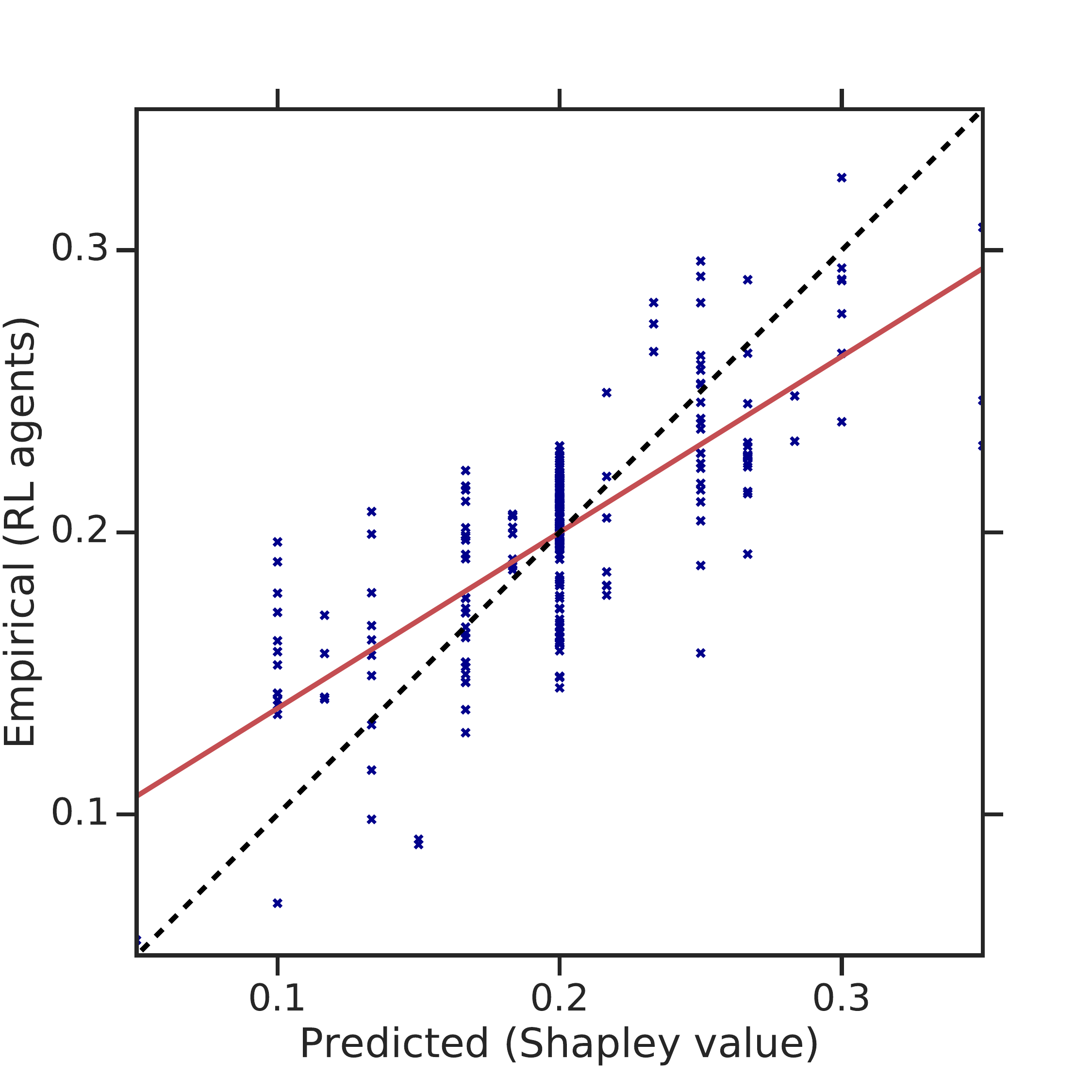}
        \includegraphics[width=0.49\textwidth]{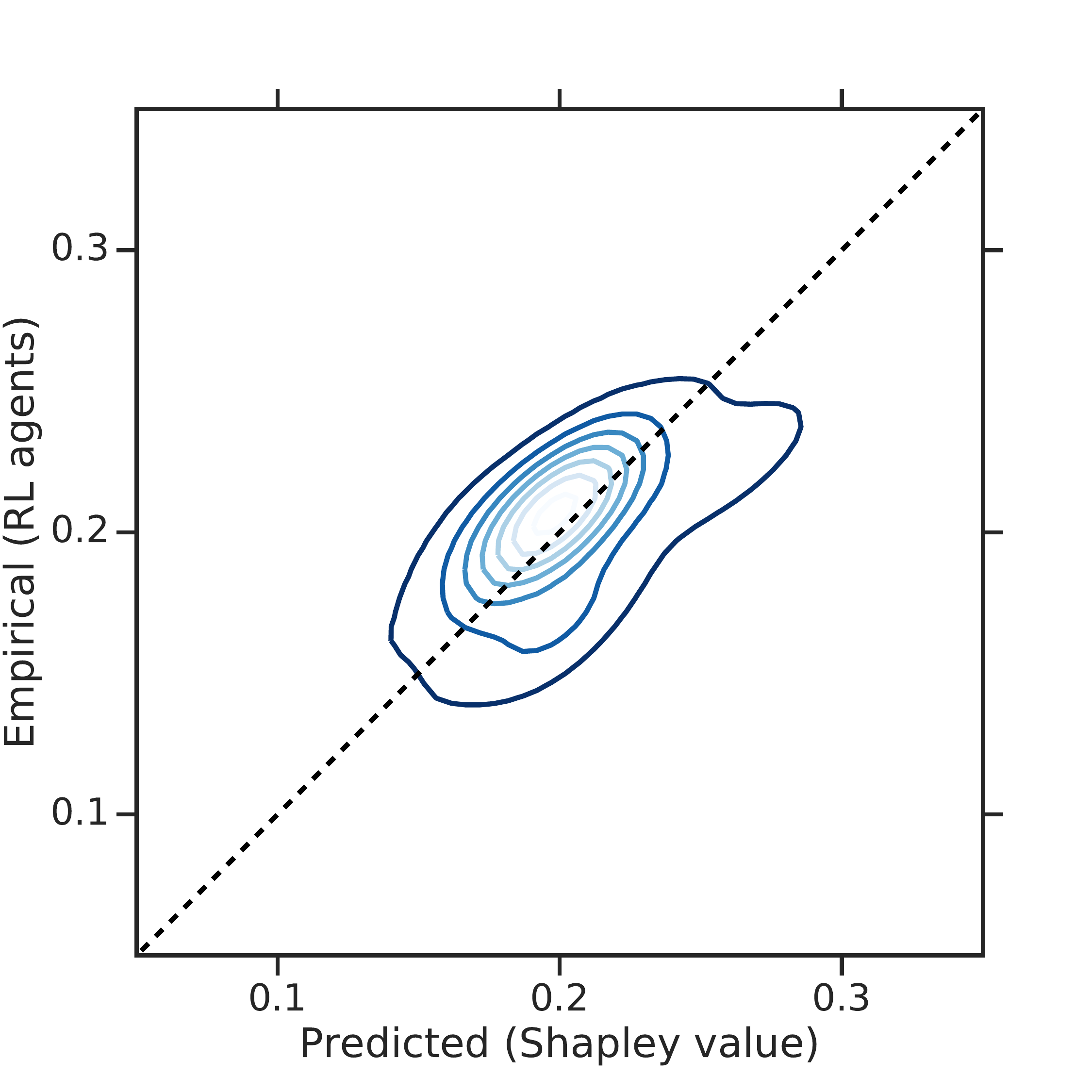}  
    \end{subfigure}
  \caption{Shapley Correspondence on Boards With Reduced Variance (Propose-Accept Environment). }
  \label{fig:shapley_reduced_var}
\end{figure}

We examine an alternative board distribution, $D'$, which does {\em not} rule-out equal-weight boards (the remaining weight sampling procedure was identical to the original distribution $D$). We generated a new train set and test set of boards from $D'$. The sampled weights under $D$ had a standard deviation of $STD_{D} = 1.65$ whereas the standard deviation of weights under $D'$ was $STD_{D'} = 1.1$. We repeated the analysis of Section~\ref{l_sect_experiment_shapley} for these reduced variance boards, resulting in Figure~\ref{fig:shapley_reduced_var}. 

Figure~\ref{fig:shapley_reduced_var} shows that the mass is much more concentrated around the area with equal Shapley values (points with $x \approx 0.2$), where the outcome achieved by RL agents is also very close to the same value (i.e. $y \approx 0.2$). As the Figure shows, this results in having a stronger correspondence between the outcome obtained by RL agents and the Shapley value. In other words, high inequality in agent power (negotiation position) results in a larger discrepancy between the Shapley value and outcomes that RL agents arrive at. 



\subsection{Influence of spatial perturbations}
\label{sec:exp_3}
As the Team Patches environment is a spatial one, we can examine how spatial aspects, that are abstracted away in the underlying cooperative game, affect negotiation outcomes. 
How does changing the spatial structure of the environment influence the ``power'' of agents? Can we reduce the share of the total reward received by the ``most powerful'' agent (with the highest weight)? We note that such factors are not taken into account in the underlying cooperative game, which only considers which agent teams are successful, as captured by the weights in the underlying game.

\begin{figure}[htbp]
\centering
\begin{subfigure}{.45\textwidth}
  \centering
  \includegraphics[width=0.75\linewidth]{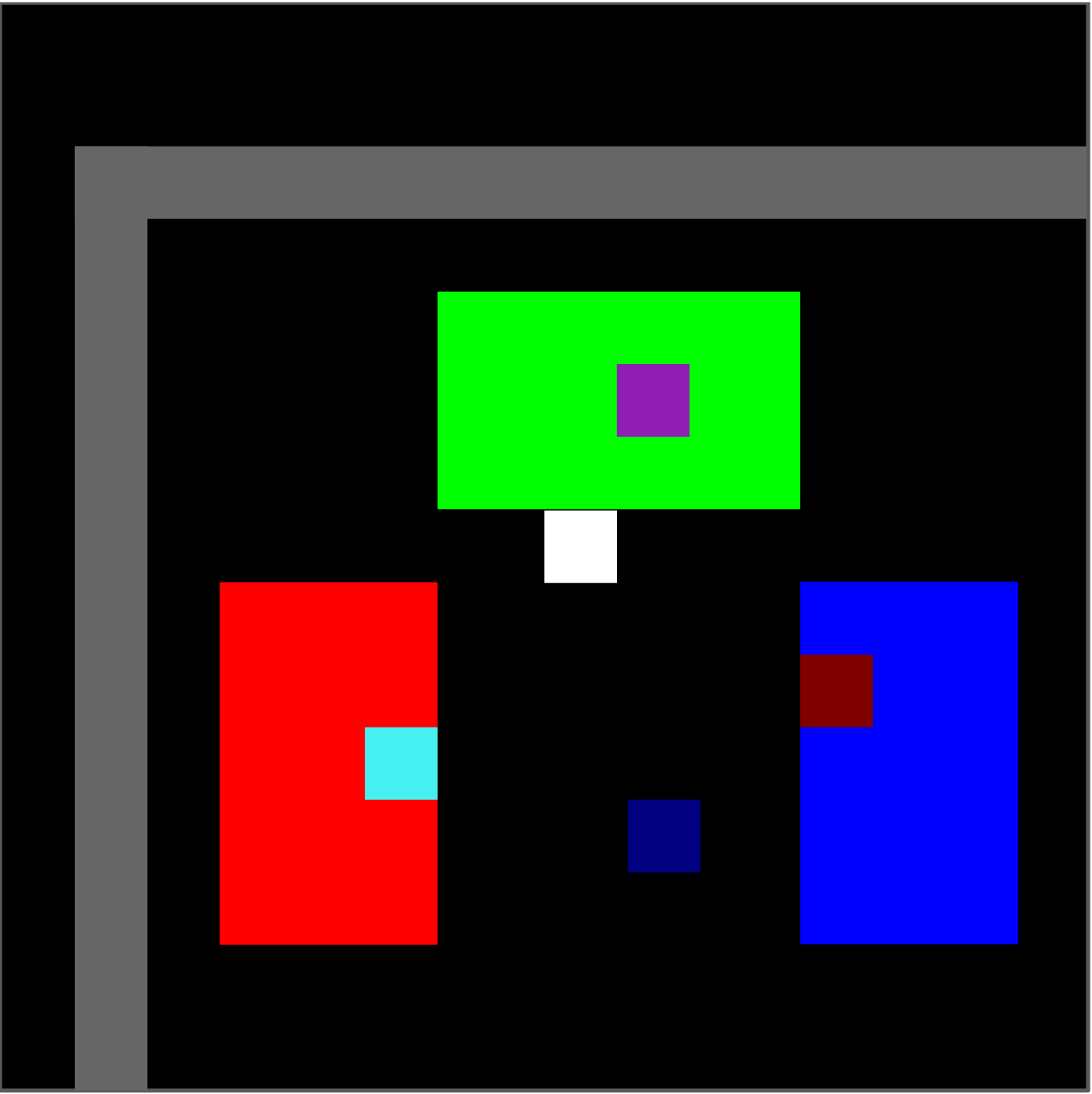}
  \caption{Example agent visual observation, from the perspective of the white agent (centered).}
  \label{fig:agent-observation}
\end{subfigure}
\hspace{2em}
\begin{subfigure}{.45\textwidth}
  \centering
  \includegraphics[width=0.75\linewidth]{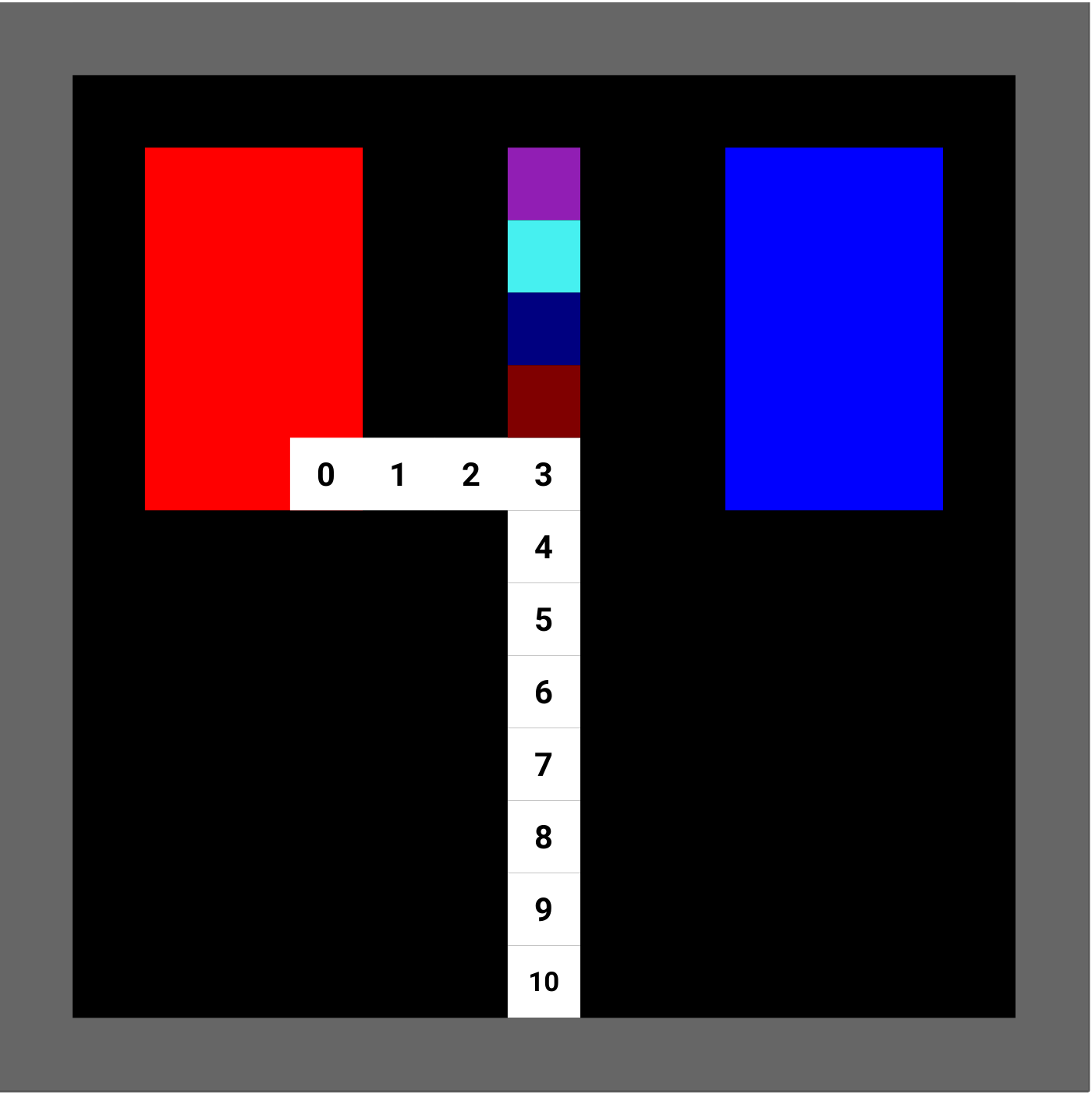}
  \caption{Visualization at the start of an episode for the spatial perturbations experiment.}
  \label{fig:spatial-perturb-visuals}
\end{subfigure}
\caption{Team Patches -- (a) Example RGB observation an agent receives in an episode. The white square is the observing agent (centered). Other agents are observed as maroon, cyan, purple, and navy, patches as red, green, and blue, and the environment's border is observed as gray. In addition to these observations, the agent also receives its index as a one-hot vector of size number of players, the weights of all agents, and their current demands. (b) Visualization of spatial perturbations where the agent with the highest weight is initialized at $0$ to $N$ squares from the nearest patch.}
\label{fig:appendix_team_patches}
\end{figure}

Our spatial Team Patches environment is a $15 \times 15$ grid-world where each entity has an assigned color. The agents observe this world from their own perspective, as shown in Figure~\ref{fig:appendix_team_patches}(a), and also observe the weights and demands of every other agent, as well as their own index. We consider two patches and vary the starting position of the maximal weight agent. We measure how the agent's share of the total reward changes as we incrementally change its starting position from directly touching a patch to being far away from the average patch location, while every other agent always starts relatively close to the nearest patch. More precisely, we change the original Team-Patches configuration in two ways:
\begin{enumerate}
    \item We set the total number of patches in the world to 2 (red and blue).
    \item We modify the starting position of the agent with the highest weight, investigating how this spatial perturbation influences their share of the total reward.
\end{enumerate}    
In Figure~\ref{fig:appendix_team_patches}(b) we visualize this with a full view of the environment where the white agent is moved between squares 0 and 10, corresponding to being $N$ steps away from the nearest patch in $L^1$ distance.

\begin{figure}[htbp]
\centering
  \includegraphics[width=0.8\textwidth]{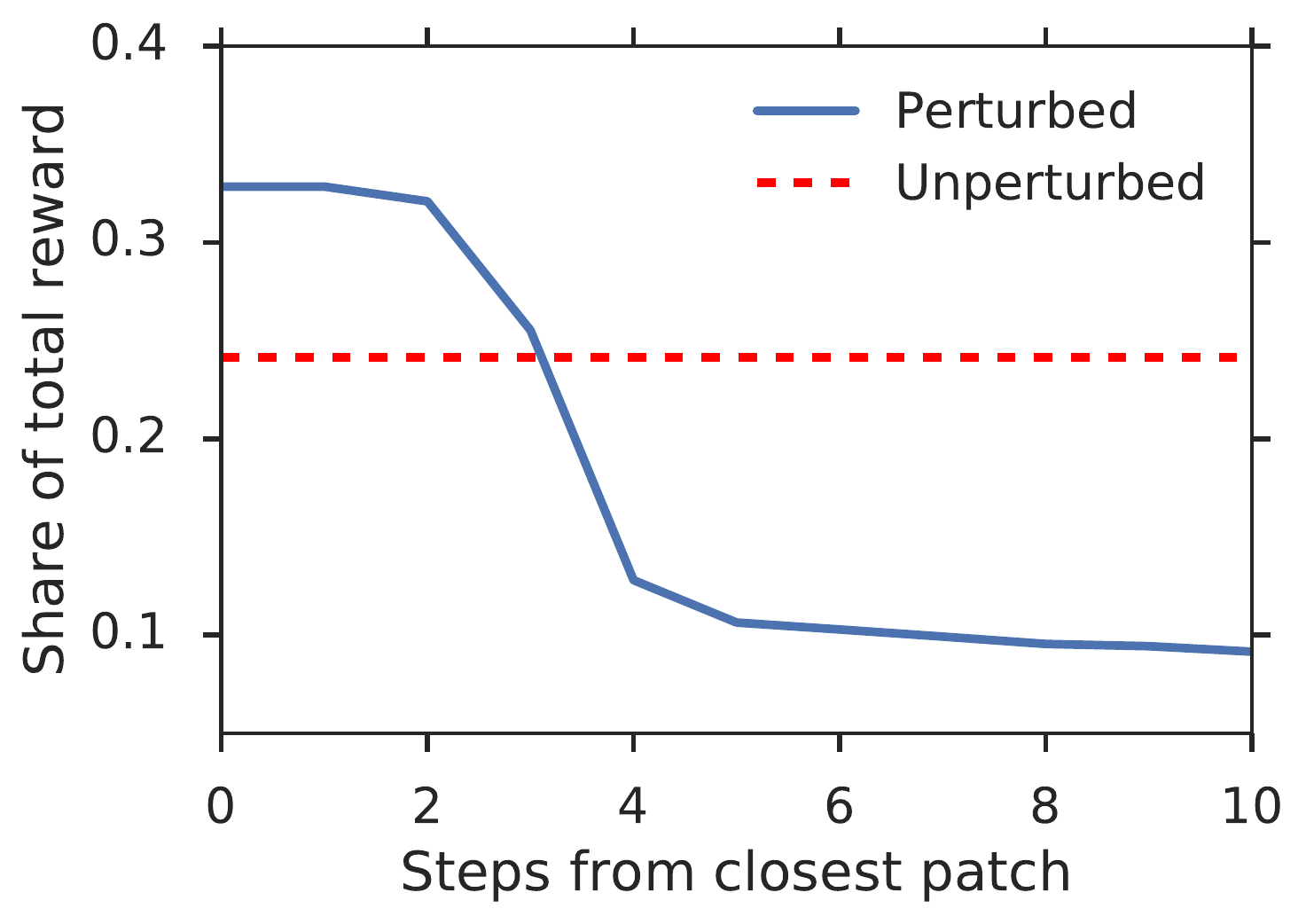}
  \caption{Influence of spatial changes (perturbed, blue) versus no changes (unperturbed, red) on the highest-weighted agent's reward.}
  \label{fig:spatial_eval}
\end{figure}

Figure~\ref{fig:spatial_eval} shows how the average share of the total reward of an agent changes as its starting position changes from being close to a patch and far from a patch (contrasting the outcome with that reached in the original, unperturbed, configuration).

%
As Figure~\ref{fig:spatial_eval} shows, when the highest-weight agent is closer to the patch than the other agents, it can start negotiating a team earlier, and due to its high weight it can demand a high reward. However, as we move the agent further away, it takes longer for the agent to reach a patch, so the other (weaker) agents have more time to form a team; thus, its share of the reward significantly drops. 
\footnote{These results also hold when shortening the number of timesteps in the episode, which effectively gives the agents less time to negotiation, hence making negotiation more difficult.}


The results of the above experiment highlight how an RL based simulation can allow us to investigate the impact of spatial factors that are abstracted away in the underlying cooperative game. Specifically, this experiment shows that giving agents ample time to negotiate gives them a negotiation edge over the competition.

\subsection{Why do RL Agents Deviate From the Shapley Value?}
\label{l_sect_shapley_distil}


Section~\ref{l_sect_experiment_shapley} shows that while for most boards our RL agents do indeed converge to an outcome that approximates the Shapley value, there is a deviation on boards where some players have particularly strong or particularly weak negotiation positions (Shapley values). 
Multiple factors may contribute to this. One potential factor relates to the \textit{representational capacity} of the RL agents; computing the Shapley value in a weighted voting game is an NP-hard problem~\cite{elkind2009computational}, so perhaps such a function cannot be easily induced with a neural network. Even if a neural network can compute the negotiation position in the underlying weighted voting game, we may still have an optimization error; RL agents optimize for their individual reward when negotiating under a specific protocol, and it might be difficult to express a policy exactly fitting the Shapley value in the negotiation protocols we've selected. Finally, the deviation might be caused by the \textit{learning dynamics} of independent RL agents; the highly non-stationary environment induced by agents learning at the same time may lead agents to agreements deviating from the cooperative game theoretic predictions.  

\begin{figure}[htbp]
\centering
    \begin{subfigure}{1.0\textwidth}
        \centering
        \includegraphics[width=0.49\textwidth]{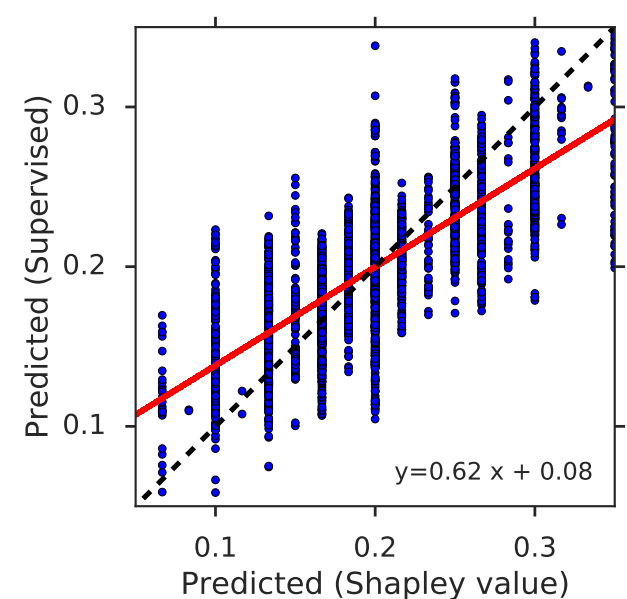}
        \includegraphics[width=0.49\textwidth]{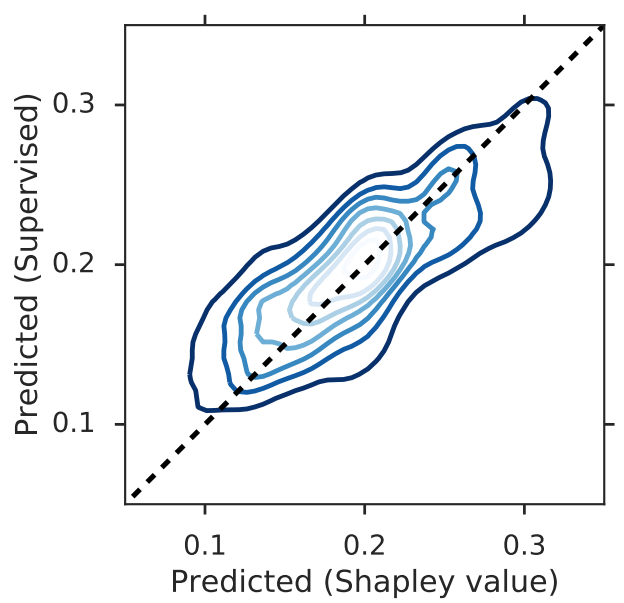}  
    \end{subfigure}
  \caption{Prediction of the Shapley values from the game weights and threshold using supervised learning.}
  \label{fig:supervised}
\end{figure}

We show that a neural network can approximate the Shapley value under supervised learning (i.e. when it does not need to consider the environment resulting from a specific negotiation protocol). We train a model to take the parameters of a weighted voting game (i.e. agent weights and the threshold), and output the Shapley values of each player. We generate $3,000$ \textit{unique} boards from the same distribution with Gaussian weights used for the experiments in Section~\ref{l_sect_experiment_shapley}, and apply a train/test partition (80\%/20\% of the data). We then fit a simple 3-layer MLP with 20 hidden units (smaller than used in the RL agents), minimizing the mean-squared error loss between the model's predictions and the Shapley values. The results of this are shown in Figure~\ref{fig:supervised}.


We find that the MLP generalizes to unseen boards, and better matches the game theoretic predictions of the Shapley value.

 
 
\subsubsection{Letting Agents Observe their Negotiation Position}
\label{l_sect_providing_shapley_observations}

In the discussion above, we suggested several potential reasons for the empirical gains of our RL agents deviating from the Shapley value. As Figure~\ref{fig:supervised} indicates, given a direct supervision signal (boards labeled with the Shapley values), a small neural net can approximate the Shapley value very well. Our RL agents have a more challenging task for two reasons: (1) They have to take into account not only their negotiating position but also protocol details. (2) Their RL supervision is weaker: they only know how successful a whole {\bf sequence} of actions was, and not the ``correct'' action they should have taken at every timestep. 

\begin{figure}[htbp]
\centering
    \begin{subfigure}{1.0\textwidth}
        \centering
        \includegraphics[width=0.49\textwidth]{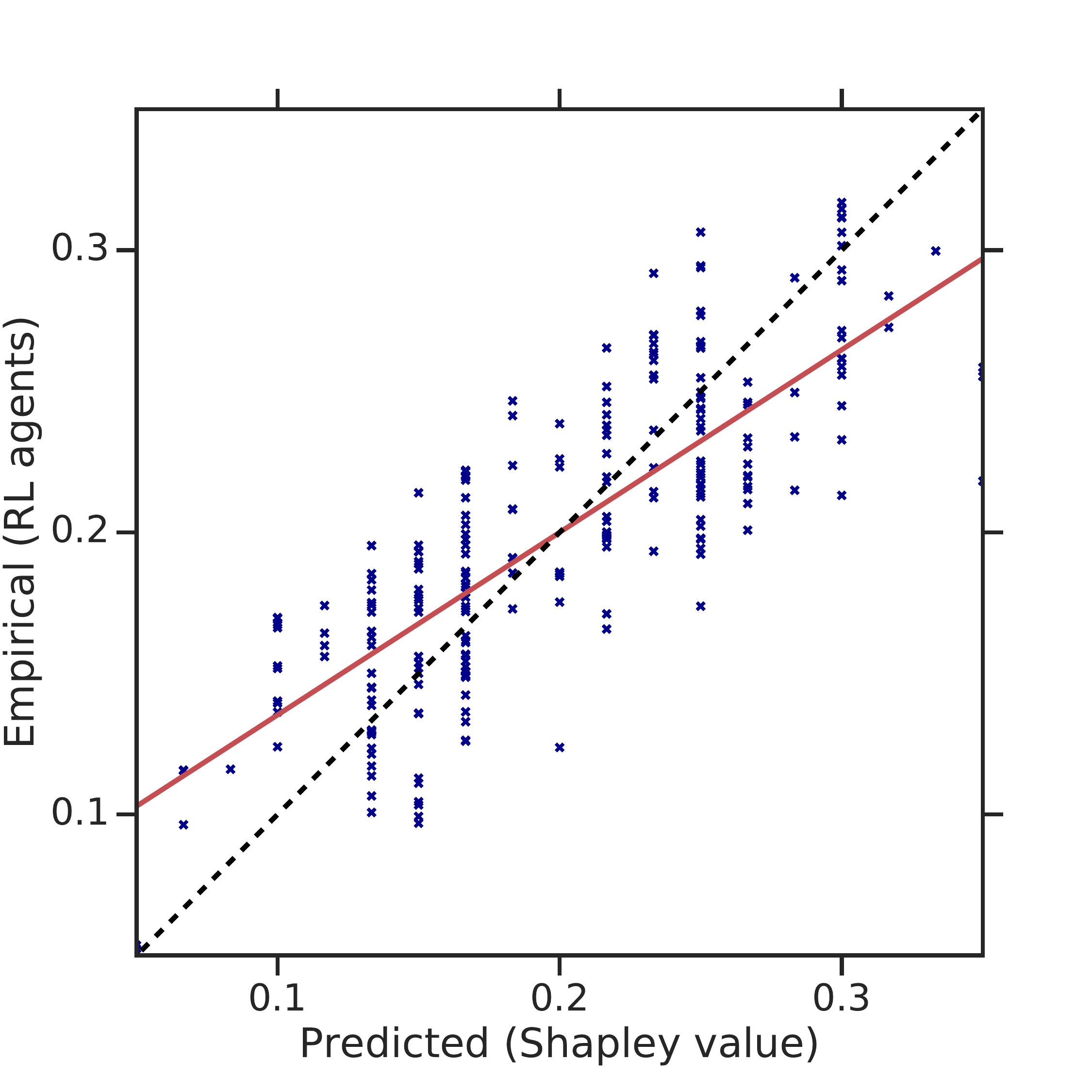}
        \includegraphics[width=0.49\textwidth]{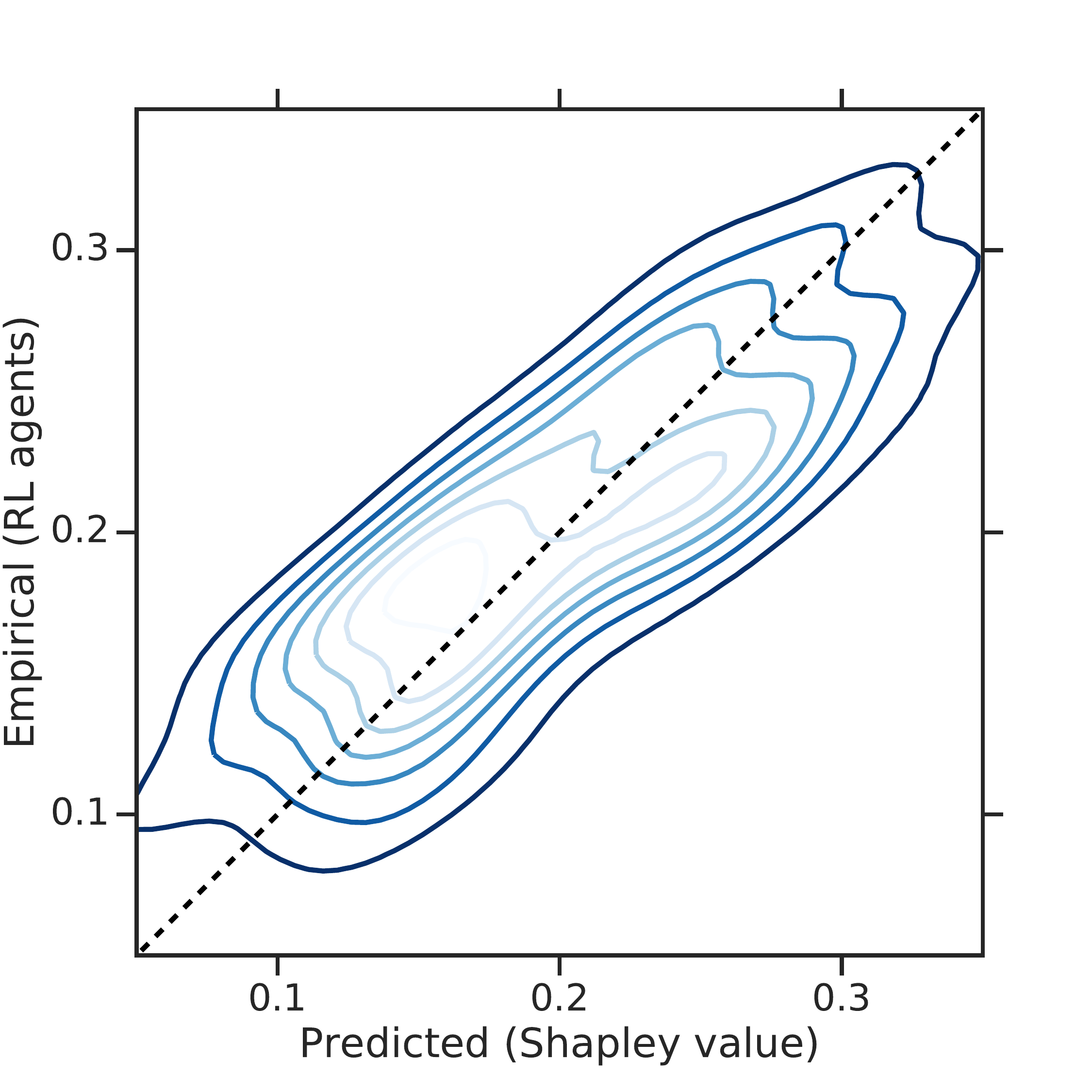}  
    \end{subfigure}
  \caption{Shapley-Aware Propose-Accept}
  \label{fig:shapley_aware}
\end{figure}

Our conclusion from Figure~\ref{fig:supervised} is that at least the basic supervised learning task can be accomplished with a small neural network, so the agent’s network has the capacity required to estimate their raw negotiating power, abstracting away protocol details. We now show that even when given the Shapley values in the underlying game, RL agents reach outcomes that may deviate from the Shapley value. We use an experimental setup that is identical to that used to produce Figure~\ref{fig:propose_accept_shapley}, except we provide the Shapley values of all the agents as a part of the observations in every state (we refer to this a Shapley-Aware Propose-Accept environment). The outcome of the experiment is shown in Figure~\ref{fig:shapley_aware}. 

Figure~\ref{fig:shapley_aware} shows the same pattern as Figure~\ref{fig:propose_accept_shapley}. The deviation from game theoretic predictions thus stems not from being unable to identify the ``high-level'' negotiation position of agents, but rather from the RL procedure we applied; RL agents attempt to maximize their own share of the gains (rather than for obtaining their ``fair'' share of the gains, as captured by the Shapley value), and are forced to deal with a the need to find strong {\em policies} taking into account the specific negotiation protocol (under a non-stationary environment, where their peers constantly adapt their behavior). 
We believe that this experiment gives some evidence that the deviation from the Shapley value does not stem from the RL agents' inability to determine their relative negotiation power; The RL agents either do not manage to find a strong policy under the specific negotiation protocol, or converge on an outcome different from the Shapley value due to having a non-stationary environment with independent learners.

\subsection{On the Relation Between the Shapley Value and Nash Equilibria in Our Negotiation Environments}
\label{l_shapley_vs_nash}

In Section~\ref{l_sect_intro} we presented our approach, which is based on investigating how agents can negotiate team formation through a {\it cooperative} game theoretic prism. The key solution concept we have applied in this work is the Shapley value, which we used as a measure of agent power in cooperative games (see Section~\ref{l_sect_appendix_power_in_politics_example}). 

A cooperative game relates to a competition between groups of players (which are referred to as ``teams'' or ``coalitions''), typically in a setting where there is an external enforcement of cooperative behavior~\cite{chalkiadakis2011computational} (for instance, such enforcement can be due to contract law). In contrast, in non-cooperative games agreements have to be ``self-enforcing'', for instance due to credible threats (or sometimes, the analysis ignores the forging of alliances altogether). In other words, the analysis of cooperative games is focused on making predictions regarding which {\it teams} will form, what {\it joint} actions the teams are likely to take, and how the payoffs would be allocated. Non-cooperative game theory, on the other hand, attempts to predict the actions that would be taken by {\it individual} players and the resulting {\it individual} utilities~\cite{osborne1994course}, typically by examining Nash equilibria~\cite{Nash1950}. 

As discussed in Section~\ref{l_game_setting}, cooperative game theory only considers the structure, strategies and payoffs of teams (captured by a cooperative game based on a characteristic function), and {\it abstracts away} the specific mechanism the individual agents use to negotiate with one another. By defining a {\it specific} protocol through which agents negotiate, including the {\it actions} that individual agents may take, we can also view the setting as a {\it non-cooperative game}, and solve it using tools such as Nash equilibrium analysis. However, the predictions of the Nash equilibrium analysis would depend on the specific interaction protocol used. 
In Section~\ref{l_sect_nego_env} we proposed two specific negotiation protocols, yielding the Propose-Accept environment~\ref{l_sect_propose_accept_env} and the Team Patches environment~\ref{l_sect_team_patches_env}. As shown in Figure~\ref{fig:overview}, though both environments may be based on the same underlying cooperative game (with the same characteristic function of a weighted voting game), they are two different Markov games. As we discuss in Section~\ref{l_sect_overview}, solving for the Nash equilibrium of even a two-player unrepeated general sum game is computationally hard, being PPAD-complete~\cite{ChenDeng06}. We have thus used a multi-agent reinforcement learning for our analysis. 

Despite the general computational hardness results, we have managed to solve for a Nash equilibrium of a simplified version of the Propose-Accept game, in the case where the time horizon of the game is fixed (i.e. there is a fixed number of rounds, rather than a certain probability of progressing to the next round). We briefly discuss this analysis, which is based on the application of a solution technique called backwards induction~\cite{osborne1994course}. We then empirically analyze our Propose-Accept environment, and show that the solutions predicted by this non-cooperative Nash equilibrium analysis are correlated with the Shapley value, though the two are not identical. 

We consider the Propose-Accept environment with a weighted voting game $[w_1, \ldots, w_n; q]$ and a total integer reward $r$, where the number of proposals is fixed to exactly $T$ rounds (where all players know the number of rounds $T$). We apply the principal of backwards induction, solving the game from the last timestep backwards towards the first round. 

Suppose that we are now in the final round. Denote by $t$ the number of {\it remaining} rounds, so we have $t=0$ remaining rounds. Suppose that a proposal has been made to player $i$, offering them a reward $x>0$. Player $i$ may accept the offer, giving them a reward of $x>0$ if all other proposees accept as well, or reject the proposal. In the case of a rejection, as this is the final round ($t=0$), the game would terminate with a reward of zero to all players. Hence, clearly the rational strategy for player $i$ is to accept this proposal (even a small reward is preferable to zero reward). Let us denote the minimal (threshold) offer for a player $i$ to accept an offer when there are $t$ remaining rounds as $a^t_i$. The minimal non-zero integer reward is 1, so we can restate the above by saying that $a^{t=0}_i=1$ for any player $i$. 
 
Now consider the proposer in this final round $t=0$. If the proposer is player $i=1$, they can choose any winning coalition (containing them), and any integral reward allocation. Given the previous argument, as this is the final round $t=0$, any player $j$ offered a reward of at least $a^{t=0}_j=1$ would accept the offer. Knowing that any player $j$ offered a reward of at least $a^{t=0}_j=1$ would take the offer, player $i=1$ would select some winning coalition $C$, give each other player $j \in C \setminus \{i\}$ the minimal reward of $a^{t=0}_j=1$ and keep $r - \sum_{j \in C \setminus \{i\}} a^t_j$ (offering any less than $a^{t=0}_j=1$ would mean $j$ would reject the offer, but there is no point offering any more, as this would come out of the proposer's pocket). This offer would be accepted by the other players (again, due to the previous argument), and as a result player $i=1$ would achieve a reward of $r - \sum_{j \in C \setminus \{i\}} a^t_j$. 
\footnote{For the final round $t=0$ we have $r - \sum_{j \in C \setminus \{i\}} a^t_j = r - |C|$, however we write the equations for a general round $t$).}

To maximize its utility $r - \sum_{j \in C \setminus \{i\}} a^t_j$, player $i$ would select the coalition $C$ that minimizes the payment to other players $\sum_{j \in C \setminus \{i\}} a^t_j$, subject to the constraint of having to choose a coalition that contains $i$ and is winning (i.e. a coalition $C$ such that $\sum_{x \in C} w_x \geq q$). Denote the winning coalitions that contain $i$ as $W_i = \{ C | i \in C, \sum_{x \in C} w_x \geq q \}$ (note $W_i$ is a set of {\it coalitions}). Thus, at time $t$, the minimal payment $i$ has to make to other players to guarantee acceptance is $g^t_i = \min_{C \in W_i} \sum_{j \in C \setminus \{i\}} a^t_j$. For a proposer $i$ at time $t$, we denote the set of coalitions that result in this minimal outgoing payment as $C^t_i = \{C \in W_i | \sum_{j \in C \setminus \{i\}} a^t_j = g^t_i \}$ (i.e. at time $t$, if player $i$ is a proposer they can select any coalition in $C^t_i$ and any of these would maximize its utility). We denote the amount that player $i$ would win when selected as the proposer at time $t$ by $d^t_i$, which as per the discussion above is $d^t_i = k - g^t_i$. 

We analyze the case where when a proposer $i$ has multiple coalitions $C^t_i$ it could propose that maximize its utility (i.e. $|C^t_i| > 1$), they choose one of these coalitions to propose uniformly at random. By proposing a coalition in $C_x \in C^t_i$, player $i$ makes a proposal that will be accepted (as each of proposee $j \in  C_x$ is offered its acceptance threshold $a^t_j$), so players outside of this proposed coalition will gain zero (as the game terminates with an accepted proposal that does not include them). Suppose $i$ is indeed the proposer at time $t$ and that $i$ has decided to propose the coalition $C_x \in C^t_i$, and consider how much a player $j$ stands to gain due to this proposal, which we denote as $S(C_x, j, i, t)$ (where $i$ is the proposer, $C_x$ is the proposed coalition, $j$ is the target player, and $t$ is the number of remaining rounds). For a player $j \notin C_x$, we have $S(C_x, j, i, t) = 0$ as $j$ is not in the proposed coalition. For a player $j \in C$, its utility depends on whether they are the proposer or one of the other players: as per the discussion above, a proposer $i$ striving to maximize its share gives each other player in the proposed coalition the minimal amount causing them to accept, and keeps the remaining reward. Hence, if $j \neq i$ we have $S(C_x, j, i, t) = a^t_i$, and if $j=i$ we have $S(C_x, j, i, t) = S(C_x, i, i, t) = d^t_i$. 

The above discussion presents multiple variables, $a^t_i$, $g^t_i$, $d^t_i$, $C^t_i$, $S(C_x, j, i, t)$ each only depending on other variables of timestep $t$. Further, given that $a^{t=0}_i = 1$ as per the discussion above, we can compute all these variables for timestep $t=0$ (i.e. zero remaining game rounds). Hence, to compute all these variables for any timestep, we need only show how the variable $a^t_i$ depends on the variables from the previous timestep $t-1$. Now consider player $i$ who rejects an offer when the game has $t>0$ remaining rounds. In this case we proceed to the next round, and have only $t-1$ remaining rounds. With probability $1/n$ player $i$ will be chosen to be the proposer, in which case they stand to gain the proposer's utility $d^{t-1}_i$. Each of the other players $j \neq i$ are also chosen with probability $1/n$, and when player $j$ is selected to be the proposer they choose a coalition $C_x$ out of the coalitions in $C^{t-1}_j$ (with each coalition $C_x$ having an equal probability of $\frac{1}{|C^{t-1}_j|}$ to be chosen), which yields player $i$ the utility $S(C_x, j, i, t-1)$. A player $i$ will accept an offer if it exceeds its expected payoff when rejecting and proceeding to the next round, and as proposals are integers, we obtain:

$$ a^t_i = 1 + \frac{1}{n} d^{t-1}_i + \sum_{j \neq i} \sum_{C_x \in C^{t-1}_j} \frac{1}{n \cdot |C^{t-1}_j|} S(C_x, j, i, t-1) $$ 

The above recursion allows us to solve for the variables for any round $t$, and thus to solve for the Nash equilibrium of the game. We now illustrate this procedure using the game $[w_1=0.4, w_2=0.4, w_3=0.2, w_4=0.2, w_5=0.2; q=1]$ with a fixed integer reward of $r=20$ and a fixed number of rounds $T=3$ (where each round includes both a propose phase and an accept/reject phase). Consider the last round $t=0$. As this is the last round, any player would accept any non-zero offer. Hence, $a^{t=0}_i = 1$ for any player $i$. If player $1$ is the proposer, they can select any winning coalition containing them, offer $a^{t=0}_i = 1$ to all other members of the coalition and keep the rest. Given the weights, we note that the set of winning coalitions containing player $1$ are $W_1= \{ \{1,2,3\}, \{1,2,4\}, \{1,2,5\}, \{1,3,4,5\} \}$. Although $\{1,3,4,5\} \in W_1$ is a winning coalition containing $1$, this coalition requires player $1$ to pay each of $\{3,4,5\}$ their threshold offer $a^{t=0}_3 = a^{t=0}_4 = a^{t=0}_5 = 1$, which results in a total payment of $\sum_{j \in \{3,4,5\}} a^{t=0}_j = 3$. In contrast, the coalition $\{ 1,2,3 \}$ is a winning coalition containing player $1$, but requires player $1$ to pay others less: $\sum_{j \in \{2,3 \}} a^{t=0}_j = 2$. Examining all the coalitions in $W_1$ we observe that the coalitions $\{ \{1,2,3\}, \{1,2,4\}, \{1,2,5\} \}$ all require player $1$ to make an equal payment of 2 to others, hence 
$C^{t=0}_1 = \{ \{1,2,3\}, \{1,2,4\}, \{1,2,5\} \}$ and $g^{t=0}_i = 2$. Hence, if player $1$ is the proposer in the final round, it is going to choose one of these coalitions in $C^{t=0}_1$ (uniformly at random), offer each of the remaining players a payment of $a^{t=0}_1 = 1$ (for all these coalitions there are two players beyond player $1$, so the total payment is $g^{t=0}_1 = 2$), and retain the remaining reward of $k-2$, so $d^{t=0}_1 = k - g^{t=0}_1 = 20 - 2 = 18$. In the case where the coalition $C_x = \{1,2,3\}$ is proposed by player $1$ (which occurs with probability $\frac{1}{|C^{t=0}_1|} = \frac{1}{3})$), player $1$ would keep $d^{t=0}_1$, and give each of $C_x \setminus \{1\} = \{ 2, 3\}$ their threshold $a^{t=0}_2 = a^{t=0}_3 =1$, and keep the reminder $d^{t=0}_1 = 18$. Thus, $S(C_x, 1, 1, 0) = 18$ and $S(C_x, 2, 1, 0) = S(C_x, 3, 1, 0) = 1$ (and for any $j \notin \{ 1, 2, 3 \}$ we have $S(C_x, j, 1, 0) = 0$). Similarly, we can compute $S(C_y, j, i, 0)$ for any winning coalition $C_y$, and any players $i,j$. Following this, for any player $i$ we can compute $a^{t=1}_i$, its minimal offer to accept for the round just before the final round (i.e. when there is $t=1$ round remaining); we can then compute the remaining variables for timestep $t=1$ ($g^{t=1}_i$, $d^{t=1}_i$, $C^{t=1}_i$, and $S(C_x, j, i, t=1)$), and so on. Eventually, using this dynamic programming approach we can compute the variables for the first round of the game $t=T-1$. 

Our propose-accept environment selects the proposer at the first timestep uniformly at random. We note that the formula for $a^{t=T-1}_i$ computes the minimal offer player $i$ should {\it accept} at time $t=T-1$ by adding one (the minimal integer increase) to their expected utility (with the expectation taken over the identity of the proposer), so player's $i$ expected utility in this game is $a^{t=T-1}_i - 1$. To determine the expected utility of a player $i$ in a game with $T$ rounds, we can examine the acceptance threshold of that player $a^{t=T-1}_i$. This is the minimal integer offer strictly exceeding the expected payoff of the player in the beginning of the round after which there are $T-1$ additional rounds (i.e. a total of $T$ rounds). The expected utility of the player in a game with $T$ rounds is thus 
$v^T_i = a^{t=T-1}_i - 1$. Applying the above computation for the game $[w_1=0.4, w_2=0.4, w_3=0.2, w_4=0.2, w_5=0.2; q=1]$ we have the following player utilities $v^{t=10}_1=v^{t=10}_2=6.29$ and $v^{t=10}_3=v^{t=10}_4=v^{t=10}_5=2.473$. In our example we have used a value of $r=20$, so $\sum_{i=1}^5 v^{t=10}_i = 20$. 
Renormalizing the values (setting $u^{t=10}_i = \frac{v^{t=10}_i}{\sum_j v^{t=10}_j})$, we get $u^{t=10}_1 = u^{t=10}_2=0.315$ and $u^{t=10}_3 = u^{t=10}_4= u^{t=10}_5 = 0.124$. The Shapley values for this game are $\phi_1 = \phi_2 = 0.3$ and $\phi_3 = \phi_4 = \phi_5 = 13.333$. Hence, in this case, the Shapley values in the underlying cooperative correlate are close to the Nash equilibrium values computed for the respective propose-accept environment (with a $T=10$ rounds). We note that, even when considering only the propose-accept protocol, the Nash equilibrium depends on the specific number of rounds chosen (i.e. when computing the backwards induction solution for a different number of rounds, we get different player utilities). 

\begin{figure}[htbp]
\centering
    \begin{subfigure}{1.0\textwidth}
        \centering
        \includegraphics[width=0.49\textwidth]{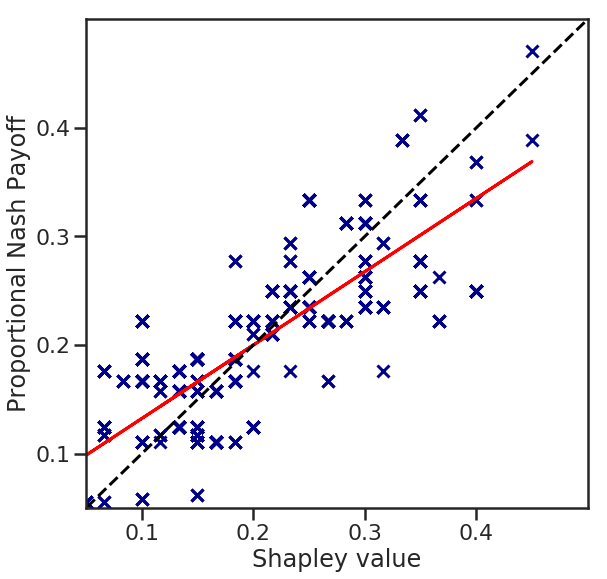}
        \includegraphics[width=0.49\textwidth]{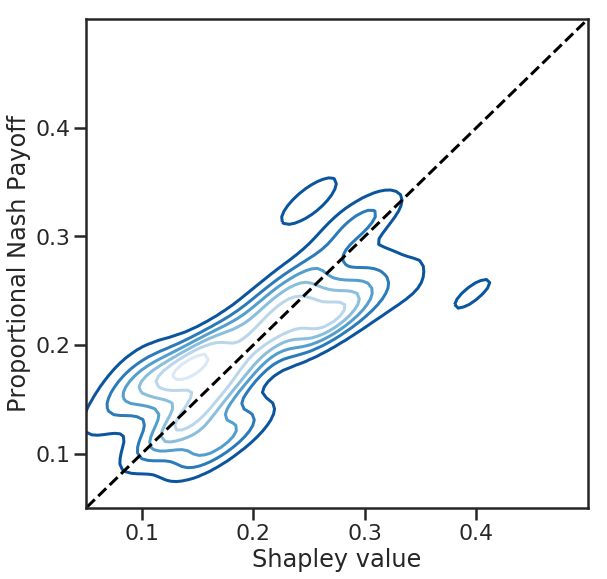}  
    \end{subfigure}
  \caption{Correlation between the Shapley value and proportional Nash equilibrium payoffs in Propose-Accept}
  \label{fig:shap-nash}
\end{figure}

We now turn to investigate the correspondence between the Shapley values of the underlying game, and the Nash equilibrium for the team formation game with the Propose-Accept protocol (as mentioned above, we can only do this in the specific case of this protocol and a fixed known number of rounds $T$). We examine the same board distribution we used for our analysis in Section~\ref{l_sect_experiment_shapley}, and apply the process discussed above to compute the Nash equilibrium for the propose-accept game with $T=10$ rounds. Figure~\ref{fig:shap-nash} shows the correlation between the Shapley value in the underlying cooperative game and the Nash equilibrium solution compute through the backwards induction process discussed above. Each point in the figure on the left reflects a single player (weight) in one of the boards used in our analysis, with the x-axis reflecting the Shapley value of the player and the y-axis reflecting the proportion of the payoff that player receives under the Nash equilibrium. The same data is shown on the right side of Figure~\ref{fig:shap-nash}, in the form of density estimation contours.

As Figure~\ref{fig:shap-nash} indicates, the Shapley values of players in the underlying cooperative game are strongly correlated with the Nash equilibrium computed using backwards induction for the corresponding Propose-Accept game. Hence, the outcomes obtained under multi-agent reinforcement learning also correlate with the Nash equilibrium (both are strongly correlated with the Shapley value). A key advantage of using the Shapley value analysis is that it is based on cooperative game theory, and relies only on the underlying cooperative game. In other words, it does not rely on a specific negotiation protocol. We emphasize that we could only solve for the Nash equilibrium in the Propose-Accept environment and not the more elaborate Team-Patches environment (in this case, the only viable option is the Shapley based analysis). Finally, we note that generally multi-agent reinforcement learning may {\it not} converge to a Nash equilibrium solution~\cite{shoham2003multi}. 

\section{Conclusions, Limitations and Future Work}
\label{l_sect_conclusions}
Team formation is an important problem for multi-agent systems, since many real-world tasks are impossible without the cooperation and coordination of multiple agents. 
The team formation problem is a hard one, since agents must both find appropriate collaborators and negotiate how to split the spoils of their team achievement. Although cooperative game theory has inspired many hand-crafted bots, these typically don't generalize across negotiation protocols, and can't easily be applied to temporally and spatially extended environments.

This work aims to contribute on multiple fronts. First, we introduced a scalable method for team-formation negotiation based on deep reinforcement learning which generalizes to new negotiation protocols and does not require human data. Second, we showed that negotiator agents derived by this method outperform simple hand-crafted bots, and produce results consistent with cooperative game theory. Further, we applied our method to spatially and temporally extended team-formation negotiation environments, where solving for the equilibrium behavior is hard. Finally, we showed that our method makes sensible predictions about the effect of spacial changes on agent behavioral and negotiation outcomes.

We used the Shapley value from {\em cooperative} game theory to examine likely outcomes in the underlying weighted voting game. The Propose-Accept environment and the Team-Patches environment are both based on this underlying game, but define the possible actions agents can take and the outcomes for various action profiles (i.e. they employ a specific negotiation protocol). Thus, they can be viewed as a {\em non-cooperative} (Markov) game. 

For such setting, one can also examine solution concepts from {\em non-cooperative} game theory, such as the Nash equilibrium~\cite{Nash1950}. While in theory one could investigate the correspondence between the outcomes reached by RL agents and the Nash equilibrium in the game, computing the equilibrium for many negotiation games is generally intractable. The Markov game formulation for either the Propose-Accept and Team-Patches games is an  $n$-player general-sum extensive-form games (with the Propose-Accept environment being an infinitely-repeated game), in which it is computationally hard to compute an equilibrium. Even the very restricted case of an unrepeated two-player general-sum game is ``hopelessly impractical to solve exactly''~\cite[Section 4.3]{SLB09}, being PPAD-complete~\cite{ChenDeng06}. While we have managed to show how to compute a Nash equilibrium in the very restricted case of the Propose-Accept protocol with a fixed time-horizon, doing so for many other protocols is very likely intractable. Hence, generally one can only apply cooperative game theoretic solutions (such as the Shapley value), and compare them to the behaviour of RL agents.

Our existing method has some limitations. The key difficulty in our framework is relying on multi-agent reinforcement learning, which has large computational costs even for very restricted domains such as our simple environments~\cite{shoham2003multi,bowling2000analysis}. This might make our solution difficult to scale-up to environments where the environment simulation itself is computationally demanding. Further, as we discussed in Section~\ref{l_sect_choice_weight_dist_convergence}, multi-agent reinforcement learning may not converge at all, and even when it does it may exhibit a different behavior from game theoretic solutions~\cite{bu2008comprehensive,lanctot2017unified}. 

This work opens up a new avenue of research applying deep learning to team-formation negotiation tasks. In particular, it would be interesting to analyze how team formation dynamics affect emergent language in reinforcement learning agents, naturally extending earlier work on cooperation and emerging communication~\cite{cao18,lazaridou16}. Indeed, it has been suggested that the human ability to negotiate and form teams was critical in the evolution of language~\cite{Thomas2018}. 

One might also consider creating tasks that interpolate between the fully cooperative game-theoretic setting and the purely non-cooperative one. Fundamentally, binding contracts are managed by dynamic institutions, whose behavior is also determined by learning. In principle, we could extend our method to this hierarchical case, perhaps along the lines of~\cite{greif2006institutions}. Perhaps such an analysis would even have implications for the social and political sciences.


\bibliography{rlnego}
\bibliographystyle{alpha}

\end{document}